\documentclass{article} % For LaTeX2e
\usepackage[T1]{fontenc}
\usepackage[preprint]{colm2025_conference}
\usepackage{amsmath}
\usepackage{microtype}
\usepackage{hyperref}
\usepackage{url}
\usepackage{booktabs}
\usepackage{cleveref}
\usepackage{lineno}
\usepackage{algorithm}
\usepackage{algorithmic}
\usepackage{subfigure}
\usepackage{subcaption}
\usepackage{caption}
\usepackage{graphicx}
\usepackage{colortbl}
\usepackage{wrapfig}
\usepackage{pgfplots}
\pgfplotsset{compat=1.17}
\usepackage{tabularx}
\definecolor{darkblue}{rgb}{0, 0, 0.5}
\hypersetup{colorlinks=true, citecolor=darkblue, linkcolor=darkblue, urlcolor=darkblue}

\title{PromptDistill: Query-based Selective Token Retention in Intermediate Layers for Efficient Large Language Model Inference}

\author{Weisheng Jin, Maojia Song, Tej Deep Pala, Yew Ken Chia \\
Singapore University of Technology and Design \\
\texttt{\{weisheng\_jin, maojia\_song, tej\_deep, yewken\_chia\}@mymail.sutd.edu.sg} \\
\And
Amir Zadeh, Chuan Li \\
Lambda Labs \\
\texttt{\{amir, c\}@lambdalabs.com} \\
\And
Soujanya Poria \\
Singapore University of Technology and Design \\
\texttt{sporia@sutd.edu.sg}
}

\begin{document}

\ifcolmsubmission
\linenumbers
\fi

\maketitle

\begin{abstract}
    As large language models (LLMs) tackle increasingly complex tasks and longer documents, their computational and memory costs during inference become a major bottleneck. To address this, we propose PromptDistill, a novel, training-free method that improves inference efficiency while preserving generation quality. PromptDistill identifies and retains the most informative tokens by leveraging attention interactions in early layers, preserving their hidden states while reducing the computational burden in later layers. This allows the model to focus on essential contextual information without fully processing all tokens. Unlike previous methods such as H2O and SnapKV, which perform compression only after processing the entire input, or GemFilter, which selects a fixed portion of the initial prompt without considering contextual dependencies, PromptDistill dynamically allocates computational resources to the most relevant tokens while maintaining a global awareness of the input. Experiments using our method and baseline approaches with base models such as LLaMA 3.1 8B Instruct, Phi 3.5 Mini Instruct, and Qwen2 7B Instruct on benchmarks including LongBench, InfBench, and Needle in a Haystack demonstrate that PromptDistill significantly improves efficiency while having minimal impact on output quality compared to the original models. With a single-stage selection strategy, PromptDistill effectively balances performance and efficiency, outperforming prior methods like GemFilter, H2O, and SnapKV due to its superior ability to retain essential information. Specifically, compared to GemFilter, PromptDistill achieves an overall $1\%$ to $5\%$ performance improvement while also offering better time efficiency. Additionally, we explore multi-stage selection, which further improves efficiency while maintaining strong generation performance. These results demonstrate the adaptability of LLM hidden representations and provide valuable insights into optimizing inference for long-context scenarios. We have made our implementations publicly available~\footnote{https://github.com/declare-lab/PromptDistill}.

\end{abstract}

\section{Introduction}
Large language models (LLMs), such as GPT, LLama, Gemini, Mistral, and Claude \citep{brown2020languagemodelsfewshotlearners, grattafiori2024llama3herdmodels, geminiteam2024gemini15unlockingmultimodal, jiang2023mistral7b, claude3} have achieved remarkable success across various natural language processing (NLP) tasks, demonstrating strong generalization and reasoning abilities. However, their widespread deployment is often hindered by the high computational and memory costs, particularly during inference with long context. As context lengths continue to grow in modern LLM applications, efficiently handling extended inputs while preserving generation quality has become a critical challenge.

Some approaches, such as MCSD and Mamba \citep{yang2024mcsd, gu2024mambalineartimesequencemodeling}, attempt to mitigate this issue by compressing internal states into a single vector. While these methods reduce processing complexity from quadratic to linear, they often suffer from performance degradation and scalability challenges, as their designs sacrifice key Transformer advantages. Other methods modify existing architectures to optimize inference efficiency, requiring extensive retraining to adapt model parameters to the new structure \citep{ijcai2024p0917, gao2024trainlongcontextlanguagemodels}. However, such approaches incur high training costs and risk compromising the original model’s capabilities.

Methods like H2O and Snap \citep{zhang2023h2oheavyhitteroracleefficient, li2024snapkvllmknowslooking} require no extra training and can be directly applied to existing models for long-context compression. However, they still process all prompt tokens through all model layers once before compression, resulting in significant computational overhead. GemFilter \citep{shi2024discoveringgemsearlylayers}, which also requires no extra training, mitigates this issue by processing only part of the layers for all prompt tokens before selecting the most important ones. It then re-runs the model using only the selected tokens for full-layer processing. While this reduces computational cost and GPU memory usage, it also discards all unselected tokens’ hidden representations, retaining only token IDs after early-layer processing. This may lead to information loss, disrupt context structures, and ultimately impact performance, especially when the selection process is imperfect or the number of retained tokens is limited.

We propose PromptDistill, a training-free method to address these limitations, achieving high efficiency while maintaining performance. First, we process a few layers for all tokens as in traditional transformers. Then, we select the most important prompt tokens based on the dot product between the query of the last prompt token and the key of each token. This selection criterion is similar to GemFilter, but instead of reprocessing the selected tokens from the first layer, we retain their hidden states after the selection layer and continue processing them through the remaining layers. The selected tokens’ hidden states already capture context information from earlier layers' self-attention, allowing us to preserve more comprehensive context and the model's original structure without additional computation.

A potential efficiency issue is that during generation, each new token attends to all prompt tokens in the initial layers, leading to high time and memory costs. To mitigate this, we introduce cache truncation, clearing the keys and values of unselected tokens in the key-value cache for layers before selection. This improves both time and memory efficiency while maintaining performance (as shown in \cref{sec:theoretical analysis} and \cref{sec:analysis}). Our method thus achieves slightly better computational efficiency than GemFilter since we avoid recomputing the selected tokens in the first few layers.

Unlike GemFilter, which can select important tokens only once in a specific layer, PromptDistill allows for multi-stage selection across layers. Early layers struggle to select tokens effectively \citep{voita-etal-2019-bottom}, but selecting more tokens in these layers can still enhance efficiency without sacrificing performance. A combination of selecting many tokens in the early layers and fewer in later layers strikes a balance between efficiency and effectiveness.

In our experiments, we apply PromptDistill to LLMs such as LLaMA-3.1, Qwen2, and Phi3 \citep{grattafiori2024llama3herdmodels, yang2024qwen2technicalreport, abdin2024phi3technicalreporthighly} and evaluate on reliable benchmarks like Longbench, Infbench, and Needle in a Haystack \citep{bai2024longbenchbilingualmultitaskbenchmark, zhang2024inftybenchextendinglongcontext, nelson2024needlehaystackmemorybased}. These extensive experiments clearly show that our method outperforms baseline methods, including full attention, GemFilter, SnapKV, and H2O \citep{shi2024discoveringgemsearlylayers, li2024snapkvllmknowslooking, zhang2023h2oheavyhitteroracleefficient}. In particular, compared to GemFilter, our method improves time efficiency and achieves better performance.

Additionally, we provide a comprehensive analysis of other aspects of our method, demonstrating how multi-stage selection improves efficiency while preserving performance.

Our contributions are summarized as:
\begin{itemize}
\setlength{\itemsep}{-1pt} % Adjust this value to reduce the gap (e.g., 0pt, 2pt, -2pt)
    \item We propose PromptDistill, a training-free, generic modification for LLMs that significantly boosts efficiency while maintaining comparable performance to full attention. It selects key tokens and retains their hidden states in intermediate layers to create an informative compression of the entire prompt.
    \item We conduct extensive zero-shot experiments with single-stage selection to evaluate PromptDistill’s ability to balance effectiveness and efficiency compared to several well-designed baselines.
    \item With multi-stage selection, PromptDistill further reduces computational costs without sacrificing performance. The experiments also provide insights into token representativeness across different layers.
\end{itemize}

\section{Method}

\subsection{Background} \label{sec:background}
To provide a clear background for our approach, PromptDistill, discussed in \cref{sec:promptdistill}, we first define the relevant preliminaries and introduce the token selection method used in our approach. For a typical LLM, let \(n\) represent the input token sequence length, \(m\) the attended token sequence length, \(d\) the hidden feature dimension, and \(i\) the index of a layer. For layer \(i\), the query matrix is \(Q^{(i)} \in \mathbb{R}^{n \times d}\), the key matrix is \(K^{(i)} \in \mathbb{R}^{m \times d}\), and the value matrix is \(V^{(i)} \in \mathbb{R}^{n \times d}\). The standard self-attention operation for each token is:
\[
\text{Self-Attn}(Q^{(i)}, K^{(i)}, V^{(i)}) = \text{Softmax}(\frac{Q^{(i)} K^{(i)\top}}{\sqrt{d}}) \cdot V^{(i)},
\]

where \(Q^{(i)} K^{(i)\top}\) computes pairwise interactions among tokens, indicating the relative importance of each token within layer \(i\). These pairwise scores are key for identifying salient segments in the input sequence.

Given the input prompt \(\mathbf{X} = (x_{1}, x_{2}, \dots, x_{n})\) (embedded in \(\mathbb{R}^{d}\) per token) for the LLM, our objective is to efficiently process it and generate high-quality responses. The GemFilter method illustrates that the LLMs can identify important tokens in early layers by attention matrices \citep{shi2024discoveringgemsearlylayers}. Based on this knowledge, the token selection method plays an important role in enabling our approach, PromptDistill, to achieve its objective.

Specifically, the token selection method identifies the top \(k\) important tokens for a given layer \(r\) using the queries and keys defined above:

\[
\text{Select}(Q^{(r)}_n, K^{(r)}, k) = \text{Top}_k \text{Argmax}_j Q^{(r)}_n K^{(r)T}_j
\]

\subsection{PromptDistill} \label{sec:promptdistill}

In \cref{sec:basic design}, we introduce the basic design of PromptDistill with single-stage selection, excluding cache truncation. This design has efficiency limitations, which are addressed by the cache truncation process described in \cref{sec:cache truncation}. The combination of both forms our proposed method. We present comprehensive experiments comparing it to multiple baselines, with results in \cref{sec:main results}. In \cref{sec:multi-stage selection}, we extend our method to multi-stage selections, with corresponding results in \cref{sec:analysis}.

\subsubsection{Basic design} \label{sec:basic design}
We outline the basic design of PromptDistill in \cref{alg:promptdistill-basic} (included in \cref{appendix: more algorithms}) and \cref{alg:promptdistill}. \cref{alg:promptdistill} shows the overall process from receiving the prompt to generating the output, while \cref{alg:promptdistill-basic} focuses on the core prompt processing and compression.

In \cref{alg:promptdistill}, after receiving the embedded prompt, \cref{alg:promptdistill-basic} processes it. We run several layers and collect the key-value cache as usual. After the $r$th layer, the token selection method in \cref{sec:background} selects the top $k$ tokens. Unlike GemFilter, we retain the hidden states of selected tokens after the $r$th layer and process the remaining layers for these tokens. This way, even unselected tokens' information is roughly captured in the selected tokens' hidden states due to prior self-attention.
\begin{wrapfigure}{r}{0.45\textwidth}  % 'r' for right, 0.45\textwidth wide
\vspace{-1.5em}  % adjust vertical space as needed
\begin{minipage}{0.45\textwidth}
\begin{algorithm}[H]
%\small
%\scriptsize
   \caption{PromptDistill}
   \label{alg:promptdistill}
    \begin{algorithmic}
        \STATE {\bfseries Input:} LLM $M$, embedded input prompt $X$, selection layer(s) $r$, number(s) of selected tokens $k$, max generated token length $T$, (optional) truncation times $tt$
        \STATE $\#$ for single-stage selection, $r,k$ will be integers
        \STATE $\#$ for multi-stage selection, $r,k$ will be lists of integers with the same length
        \STATE Generation $\gets$ [ ]
        \STATE new\_token, Cache $\gets$ LLM-forward-PromptDistill($M,X,r,k,\text{(optional)}tt$)
        \STATE Generation$_0$ $\gets$ new\_token
        \FOR{$t \gets 1 ... T-1$}
            \STATE new\_token, Cache $\gets$ LLM-forward(M, new\_token, Cache)
            \STATE Generation$_t\gets$ new\_token
        \ENDFOR
        \STATE return Generation
    \end{algorithmic}
\end{algorithm}
\end{minipage}
\vspace{-4pt}
\end{wrapfigure}
After \cref{alg:promptdistill-basic}, the first generated token and previous cache are fed into the normal LLM forward process to generate the remaining tokens. Our approach differs from typical LLMs in that the key-value cache is smaller, as only selected tokens are processed in layers after the $r$th layer.
\subsubsection{Cache truncation} \label{sec:cache truncation}
% \begin{algorithm}[tbh]
% \small
% \scriptsize
%    \caption{LLM-forward-PromptDistill (single-stage selection, with cache truncation)}
%    \label{alg:promptdistill-truncation}
%     \begin{algorithmic}
%         \STATE {\bfseries Input:} LLM $M_{0:R}$, embedded input prompt $X$, selection layer $r$, number of selected tokens $k$
%         \STATE Cache $\gets$ [ ]
%         \FOR{$i \gets 0 ... R$}
%             \STATE $X, Q^{(i)}_n, K^{(i)}, V^{(i)}\gets M_i$($X$)
%             \STATE Cache$_i \gets K^{(i)}, V^{(i)}$ 
%             \IF{$i$  equals  $r$}
%                 \STATE indices $\gets$ Select$(Q^{(i)}_n, K^{(i)}, k)$
%                 \STATE $X \gets X_{\text{indices}}$ 
%                 \FOR{$j \gets 0 ... i$}
%                     \STATE $K^{(j)}, V^{(j)} \gets $Cache$_j$ 
%                     \STATE $K^{(j)} \gets K^{(j)}_{\text{indices}}$
%                     \STATE $V^{(j)} \gets V^{(j)}_{\text{indices}}$
%                     \STATE Cache$_j \gets K^{(j)}, V^{(j)}$ 
%                 \ENDFOR
%             \ENDIF
%         \ENDFOR
%         \STATE return generate($X$), Cache
%     \end{algorithmic}
% \end{algorithm}

\begin{wrapfigure}{r}{0.45\textwidth}  % 'r' for right, 0.45\textwidth wide
\vspace{-1.5em}  % adjust vertical space as needed
\begin{minipage}{0.45\textwidth}
\begin{algorithm}[H]
%\scriptsize
\caption{LLM-forward-PromptDistill (with cache truncation)}
\label{alg:promptdistill-truncation}
\begin{algorithmic}
        \STATE {\bfseries Input:} LLM $M_{0:R}$, embedded input prompt $X$, selection layer $r$, number of selected tokens $k$
        \STATE Cache $\gets$ [ ]
        \FOR{$i \gets 0 ... R$}
            \STATE $X, Q^{(i)}_n, K^{(i)}, V^{(i)}\gets M_i$($X$)
            \STATE Cache$_i \gets K^{(i)}, V^{(i)}$ 
            \IF{$i$  equals  $r$}
                \STATE indices $\gets$ Select$(Q^{(i)}_n, K^{(i)}, k)$
                \STATE $X \gets X_{\text{indices}}$ 
                \FOR{$j \gets 0 ... i$}
                    \STATE $K^{(j)}, V^{(j)} \gets $Cache$_j$ 
                    \STATE $K^{(j)} \gets K^{(j)}_{\text{indices}}$
                    \STATE $V^{(j)} \gets V^{(j)}_{\text{indices}}$
                    \STATE Cache$_j \gets K^{(j)}, V^{(j)}$ 
                \ENDFOR
            \ENDIF
        \ENDFOR
        \STATE return generate($X$), Cache
\end{algorithmic}
\end{algorithm}
\end{minipage}
\vspace{-1em}  % adjust if you want to tighten up
\end{wrapfigure}
Although our basic design is significantly more efficient than typical LLMs, which process all tokens across all layers and retain the full key-value cache for generation, it still has efficiency limitations. Specifically, each newly generated token continues to attend to all prompt tokens in the initial layers, leading to unnecessary computational overhead (further discussed in \cref{sec:theoretical analysis}). To address this, we introduce a cache truncation mechanism to further reduce computation costs.

The forward function of PromptDistill with cache truncation is detailed in \cref{alg:promptdistill-truncation}. Compared to the basic design, after selecting important tokens and discarding the hidden states of unselected ones, we further prune the key-value cache by keeping only entries corresponding to the selected tokens across all processed layers. This significantly reduces both computation time and GPU memory usage, making our method even more time-efficient than GemFilter.

Crucially, our experiments show that cache truncation does not degrade performance (see \cref{sec:analysis}). Therefore, we adopt PromptDistill with cache truncation as our proposed method.

\subsubsection{Multi-stage selection} \label{sec:multi-stage selection}
Unlike GemFilter, which can only select tokens at a single layer, our approach allows multiple selection stages. The key insight behind multi-stage selection is that LLMs have limited ability to identify and compress important tokens in very early layers. A single selection in a mid-layer with a small number of tokens requires processing all tokens in many layers before selection, leading to high computational and memory costs. Instead, selecting a large number of tokens in early layers and refining the selection in later layers provides a better balance between efficiency and effectiveness.

\Cref{alg:promptdistill-multi} (included in \cref{appendix: more algorithms}) integrates multi-stage selection with cache truncation. Here the selection layer $r$ becomes a list of layers, and $k$ corresponds to the number of selected tokens at each stage. Unlike single-stage selection, where the cache is truncated to match the final selection size, multi-stage selection introduces a user-defined parameter $tt$, representing the number of truncation steps. This ensures that the final cache size matches $k[tt-1]$. 

The model processes layers normally until reaching a selection layer, where it retains only the selected tokens’ hidden states. If this is the $p$-th selection step ($p$ starting from $0$) and $p<tt$, truncation is applied to remove unselected tokens from all completed layers. The process repeats until all layers are completed, allowing for flexible cache structuring for further token generation.

To maintain consistency in baseline comparisons, we primarily evaluate the single-stage version of PromptDistill. However, we showcase the effectiveness of multi-stage selection in our analysis.

\subsection{Theoretical Analysis} \label{sec:theoretical analysis}
\begin{table}[t]
  \centering
  %\small
  \caption{Complexity Analysis. Here $m$ is the total layer number of LLM, $h$ is the number of attention heads, $n$ is prompt token length, $d$ is the hidden dimension of each head, $r$ is the index of layer for token selection in GemFilter and PromptDistill, $k$ is the number of selected tokens (for GemFilter and PromptDistill) and the cache size for generation (for SnapKV, H2O, GemFilter and PromptDistill), $t$ is the number of generated tokens, $w$ is the GPU memory used for model parameters in each layer, $n \geq max\{d, k, t\}$}
  \label{Complexity Analysis}
  \resizebox{\linewidth}{!}{
  \begin{tabular}{lcccccc}
  \toprule
   & \textbf{Stage} & \textbf{AllKV} & \textbf{SnapKV and H2O} & \textbf{GemFilter} & \textbf{PromptDistill (proposed)} & \textbf{PromptDistill (no truncation)}\\
  \midrule
  Time& Prompt & $\Theta(mhn^2d)$ & $\Theta(mhn^2d)$ & $\Theta(rhn^2d)$ & $\Theta(rhn^2d)$ & $\Theta(rhn^2d)$\\
   & Generation & $\Theta(mh(nt+t^2)d)$ & $\Theta(mh(kt+t^2)d)$ & $\Theta(mh(k^2+t^2)d)$ & $\Theta(mh(k^2+t^2)d-rhk^2d)$ & $\Theta(mh(k^2+t^2)d+rh(nt-kt-k^2)d)$\\
  \midrule
  GPU& Prompt & $mw + 2mhnd$ & $mw + 2hnd + 2mhkd$ & $rw + 2hnd$ & $rw + 2rhnd$ & $rw + 2rhnd$\\
   & Generation & $mw + 2mh(n + t)d$ & $mw + 2mh(k + t)d$ & $mw + 2mh(k + t)d$ & $mw + 2mh(k + t)d$ & $mw+2mh(k+t)d+2rh(n-k)d$\\
  \bottomrule
  \end{tabular}
  }
\end{table}
In \cref{Complexity Analysis}, we provide a theoretical comparison of computation time and GPU memory usage between our method and various baselines. Variable definitions are provided in the table caption. To ensure consistency, we analyze PromptDistill with single-stage selection and cache truncation. "AllKV" refers to the original LLM using full attention. Baseline values are taken from GemFilter’s paper \citep{shi2024discoveringgemsearlylayers}, with our method’s values added for comparison.

Inference efficiency is analyzed in two parts: prompt computation and token generation. For GemFilter, all $n$ tokens are processed for $r$ layers to select $k$ important tokens, which are then used for rerunning the full $m$ layers to generate tokens. They put the part of processing all tokens in $r$ layers in the prompt computation stage and the remaining, including processing the selected prompt, in the generation stage. Following this structure, we also place our computation for the first $r$ layers in the prompt computation stage, while the subsequent processing of selected tokens and token generation falls under the generation stage.

For the time cost, in the first stage, both PromptDistill and GemFilter only need to run $r$ layers for all tokens, while other methods run all layers for all tokens. This gives PromptDistill and GemFilter the same time cost in this stage, both better than other methods. In the second stage, while GemFilter runs selected prompt tokens for $\Theta(mhk^2d)$, PromptDistill only processes the remaining $m-r$ layers for the selected tokens' hidden states, requiring $\Theta((m-r)hk^2d)$. Both methods then generate $t$ tokens with a key-value cache size of $k$, taking $\Theta(mh(kt+t^2)d)$. Thus, GemFilter's total time cost in the second stage is $\Theta(mhk^2d+mh(kt+t^2)d)=\Theta(mh(k^2+t^2)d)$, while PromptDistill's is $\Theta((m-r)hk^2d+mh(kt+t^2)d)=\Theta(mh(k^2+t^2)d-rhk^2d)$. Since GemFilter is already more efficient than AllKV, SnapKV, and H2O as shown in its paper, and PromptDistill further improves generation time efficiency over GemFilter, PromptDistill achieves the best overall time efficiency.

For GPU memory usage, in the first stage, we require only $r$ layers of model weights, which is $rw$, the same as GemFilter. Since all keys and values must be stored before selecting tokens, memory usage is $2rhnd$, proportional to the number of layers before selection. In the second stage, we retain the full model weights and store up to $k+t$ tokens' keys and values for all layers, the same as GemFilter, SnapKV, and H2O.

The efficiency of the basic design without cache truncation is also shown in \cref{Complexity Analysis} for comparison. In stage one, the time cost is the same as our proposed method since cache truncation occurs after selection. In stage two, while the basic design also requires $\Theta((m-r)hk^2d)$ for processing selected tokens, it differs in key-value cache handling. Since all prompt tokens' keys and values are kept in the first $r$ layers, the generation step takes $\Theta(rh((n+t)t)d+(m-r)h((k+t)t)d)=\Theta(rh(nt+t^2)d+(m-r)h(kt+t^2)d)=\Theta(rh((n-k)t)d+mh(kt+t^2)d)$. The total time cost then becomes $\Theta((m-r)hk^2d+rh((n-k)t)d+mh(kt+t^2)d)=\Theta(mh(k^2+t^2)d+rh((n-k)t)d-rhk^2d)=\Theta(mh(k^2+t^2)d+rh(nt-kt-k^2)d)$. For GPU memory in stage two, since the cache in the first $r$ layers is not truncated, memory usage is $mw+2rh(n+t)d+2(m-r)h(k+t)d=mw+2mh(k+t)d+2rh(n-k)d$. This confirms that cache truncation is essential for efficiency.

For multi-stage selection, different configurations are possible. As an example, consider selecting $k'$ tokens at layer $r'$ and then $k$ tokens at layer $r$ with $r>r',k'>k$, finally truncating the cache to $k$ tokens. The time cost in stage one is $\Theta(r'hn^2d+(r-r')hk'^2d)$. Since a first truncation occurs after the initial selection, GPU memory usage in stage one is $max\{r'w+2r'hnd,rw+2rhk'd\}$. After stage one, the cache size is the same as in single-stage selection, so stage two has identical time and memory costs. This shows that multi-stage selection can significantly improve efficiency with well-chosen selection layers and token counts.
\section{Experiments}

\subsection{Experimental Setup}
For our main experiments, we use PromptDistill with single-stage selection and cache truncation. Since compression methods primarily improve efficiency for long inputs, we evaluate on three authoritative long-context benchmarks: LongBench, InfBench, and Needle in a Haystack \citep{bai2024longbenchbilingualmultitaskbenchmark, zhang2024inftybenchextendinglongcontext, nelson2024needlehaystackmemorybased}.

For base models, we select Llama 3.1 8B Instruct (supporting 128000 input length), Phi 3.5 Mini Instruct (supporting 128000 input length), and Qwen2 7B Instruct (supporting 32768 input length) \citep{grattafiori2024llama3herdmodels, abdin2024phi3technicalreporthighly, yang2024qwen2technicalreport}. Baseline methods include full attention (original model), SnapKV, H2O, and GemFilter \citep{li2024snapkvllmknowslooking, zhang2023h2oheavyhitteroracleefficient, shi2024discoveringgemsearlylayers}. For both PromptDistill and GemFilter \citep{shi2024discoveringgemsearlylayers}, a selection layer must be specified. We use layer 13 (out of 32) for Llama 3.1, layer 19 (out of 32) for Phi 3.5, and layer 16 (out of 28) for Qwen2. The choices for Llama 3.1 and Phi 3.5 follow GemFilter's settings for consistency, while Qwen2’s selection layer is chosen based on GemFilter’s performance across layers, which will be further discussed in \cref{sec:analysis}. In \cref{sec:analysis}, we conduct ablation studies and further exploration, testing various PromptDistill settings, including multi-stage selection and cache truncation removal.

\subsection{Main Results}\label{sec:main results}
\subsubsection{Long Bench}\label{sec:longbench}
\begin{table}[t]
  \centering
  %\small
  \caption{Evaluation Results for Longbench (PromptDistill's difference from GemFilter is labeled)}
  \label{table:longbench}
  \resizebox{\linewidth}{!}{
  \begin{tabular}{l*{15}{c}}
  \toprule
  \textbf{Method} & \textbf{NrtvQA} & \textbf{qasper} & \textbf{MF-en} & \textbf{hotpotqa} & \textbf{2wikimqa} & \textbf{musique} & \textbf{gov\_report} & \textbf{qmsum} & \textbf{multi\_news} & \textbf{trec} & \textbf{triviaqa} & \textbf{samsum} & \textbf{PCount} & \textbf{PRe} & \textbf{Mean}\\
  \midrule
  \rowcolor[gray]{.9}&\multicolumn{15}{c}{\textbf{LLama 3.1 8B Instruct}}\\
  \midrule
  %&&&&&&&\textbf{LLama 3.1 8B Instruct}\\
  All KV & 32.02 & 13.04 & 27.34 & 16.23 & 16.05 & 11.22 & 34.52 & 23.41 & 26.89 & 73.0 & 91.64 & 43.8 & 7.16 & 97.73 & 36.72\\
  H2O-4096 & 22.94 & 12.61 & 26.48 & 16.63 & 15.81 & 10.14 & 33.51 & 23.47 & 26.81 & 69.0 & 91.15 & 43.97 & 6.66 & 71.67 & 33.63\\
  \midrule
  SnapKV-1024 & 31.98 & 11.17 & 25.33 & 14.81 & 15.73 & 10.69 & 26.95 & 22.89 & 25.86 & 67.5 & 91.89 & 42.85 & 7.67 & 98.16 & 35.25\\
  GemFilter-1024 & 20.71 & 11.0 & 29.28 & 19.12 & 17.01 & 13.01 & 30.37 & 21.75 & 25.17 & 63.0 & 90.7 & 42.5 & 7.15 & 92.22 & 34.50\\
  PromptDistill-1024 & 24.24 & 11.36 & 28.01 & 16.59 & 14.92 & 11.7 & 29.68 & 22.29 & 25.77 & 73.5 & 91.65 & 43.61 & 5.87 & 95.8 & 35.36(+0.86)\\
  \midrule
  SnapKV-2048 & 31.45 & 11.94 & 26.24 & 15.73 & 16.03 & 11.66 & 29.64 & 23.24 & 26.44 & 69.5 & 91.48 & 42.68 & 7.21 & 98.03 & 35.80\\
  GemFilter-2048 & 24.36 & 12.63 & 25.39 & 19.58 & 17.03 & 14.11 & 33.15 & 22.31 & 26.49 & 69.5 & 91.59 & 42.64 & 4.61 & 98.75 & 35.87\\
  PromptDistill-2048 & 28.24 & 12.86 & 26.75 & 17.47 & 15.88 & 12.46 & 32.06 & 23 & 26.86 & 73.5 & 91.54 & 43.22 & 5.62 & 97.79 & 36.23(+0.36)\\
  \midrule
  SnapKV-4096 & 32.13 & 13.12 & 27.38 & 16.11 & 16.08 & 11.6 & 32.39 & 23.47 & 26.76 & 71.5 & 91.64 & 43.46 & 7.33 & 97.24 & 36.44\\
  GemFilter-4096 & 25.66 & 12.95 & 27.38 & 17.76 & 15.6 & 12.02 & 34.17 & 23.25 & 26.87 & 70.0 & 92.36 & 43.34 & 5.96 & 98.0 & 36.09\\
  PromptDistill-4096 & 30.2 & 13.24 & 26.91 & 16.41 & 16.12 & 11.6 & 34.2 & 23.08 & 26.86 & 72.5 & 91.47 & 44.17 & 5.24 & 97.22 & 36.37(+0.28)\\
  \midrule
\rowcolor[gray]{.9}&\multicolumn{15}{c}{\textbf{{Phi 3.5 Mini 3.8B Instruct}}}\\
  \midrule
  All KV & 27.51 & 17.23 & 35.63 & 21.7 & 25.7 & 11.68 & 34.14 & 23.17 & 24.95 & 71.5 & 87.37 & 13.08 & 7.17 & 83.85 & 34.62\\
  H2O-4096 & 19.74 & 16.23 & 34.17 & 21.02 & 23.05 & 10.49 & 33.42 & 21.95 & 24.95 & 67.5 & 86.13 & 16.71 & 1.55 & 47.46 & 30.31\\
  \midrule
  SnapKV-1024 & 24.31 & 16.03 & 34.93 & 20.72 & 26.02 & 13.74 & 28.27 & 22.03 & 24.02 & 67.5 & 87.71 & 14.57 & 6.08 & 85.6 & 33.68\\
  GemFilter-1024 & 16.57 & 18.29 & 35.91 & 24.22 & 26.1 & 9.7 & 30.29 & 18.96 & 23.64 & 64.5 & 85.85 & 23.02 & 0.2 & 81.12 & 32.74\\
  PromptDistill-1024 & 23.42 & 21.79 & 33.69 & 23.46 & 24.07 & 12.87 & 30.39 & 22.07 & 24.14 & 68.5 & 87.51 & 13.56 & 3.28 & 83.58 & 33.74(+1)\\
  \midrule
  SnapKV-2048 & 26.41 & 16.59 & 36.99 & 21.8 & 26.07 & 12.57 & 30.88 & 22.37 & 24.51 & 69.5 & 87.54 & 13.13 & 6.57 & 83.92 & 34.20\\
  GemFilter-2048 & 19.63 & 14.84 & 35.99 & 21.38 & 19.72 & 10.13 & 32.39 & 21.24 & 24.71 & 65.0 & 86.49 & 20.47 & 2.17 & 69.5 & 31.69\\
  PromptDistill-2048 & 23.85 & 17.81 & 35.98 & 21.57 & 24.78 & 11.13 & 32.79 & 22.22 & 24.7 & 69.5 & 87.42 & 13.5 & 4.43 & 82.75 & 33.75(+2.06)\\
  \midrule
  SnapKV-4096 & 27.25 & 17.42 & 36.9 & 21.37 & 25.42 & 12.55 & 32.9 & 22.6 & 24.87 & 70.5 & 87.45 & 13.28 & 6.81 & 84.04 & 34.53\\
  GemFilter-4096 & 20.95 & 19.98 & 35.22 & 28.82 & 28.21 & 13.98 & 34.2 & 22.45 & 25.08 & 64.5 & 85.86 & 18.68 & 3.43 & 65.56 & 33.35\\
  PromptDistill-4096 & 24.67 & 16.5 & 35.88 & 20.54 & 25.58 & 12.47 & 34 & 22.51 & 24.97 & 71 & 87.37 & 13.57 & 5.8 & 82.97 & 34.13(+0.78)\\
  \midrule
    \rowcolor[gray]{.9}&\multicolumn{15}{c}{\textbf{{Qwen2 7B Instruct}}}\\
  \midrule
  All KV & 24.59 & 44.4 & 45.25 & 12.99 & 12.68 & 8.9 & 37.64 & 24.68 & 27.32 & 78.5 & 88.77 & 46.57 & 6 & 74.5 & 38.06\\
  H2O-4096 & 8.98 & 15.89 & 16.64 & 21.38 & 22.42 & 8.81 & 0.02 & 0.02 & 0 & 19 & 39.8 & 0 & 8 & 56.62 & 15.54\\
  \midrule
  SnapKV-1024 & 24.48 & 42.66 & 45.41 & 12.76 & 12.33 & 8.46 & 28.97 & 24.23 & 26.43 & 71 & 88.47 & 45.89 & 6.67 & 73.17 & 36.50\\
  GemFilter-1024 & 22.8 & 38.78 & 43.34 & 12.8 & 14.49 & 9.81 & 27.85 & 21.03 & 26.04 & 74.5 & 88.55 & 43.21 & 6.88 & 46.17 & 34.02\\
  PromptDistill-1024 & 24.35 & 38.54 & 43.36 & 13.33 & 13.3 & 9.06 & 29.25 & 21.54 & 26.35 & 74.5 & 89.44 & 44.36 & 11 & 71.62 & 36.43(+2.41)\\
  \midrule
  SnapKV-2048 & 24.44 & 44.69 & 45.05 & 13.06 & 12.16 & 8.72 & 32.45 & 24.35 & 27.1 & 76 & 88.77 & 46.06 & 6.67 & 74.83 & 37.45\\
  GemFilter-2048 & 24.01 & 43.68 & 44.27 & 11.77 & 13.44 & 10.03 & 31.49 & 22.3 & 27.14 & 78.5 & 88.43 & 43.58 & 5.88 & 55.7 & 35.73\\
  PromptDistill-2048 & 25.65 & 41.99 & 44.73 & 13.07 & 13.53 & 8.01 & 33.11 & 22.47 & 27.28 & 77.5 & 88.99 & 46.55 & 8.67 & 73.67 & 37.52(+1.79)\\
  \midrule
  SnapKV-4096 & 24.87 & 44.36 & 44.14 & 13.32 & 12.39 & 8.67 & 35.06 & 24.83 & 27.3 & 77.5 & 88.77 & 46.45 & 6.75 & 75.17 & 37.83\\
  GemFilter-4096 & 25.49 & 46.62 & 43.38 & 12.09 & 13.09 & 9.03 & 33.85 & 23.28 & 27.51 & 79 & 89.49 & 44.85 & 4.92 & 64.75 & 36.95\\
  PromptDistill-4096 & 25.76 & 44.99 & 44.98 & 12.95 & 12.92 & 8.66 & 35.65 & 23.56 & 27.43 & 77.5 & 89.07 & 46.24 & 7.33 & 76.04 & 38.08(+1.13)\\
  \bottomrule
  \end{tabular}
  }
\end{table}

We first apply our method, along with all baseline approaches, to three base models on LongBench, a benchmark comprising various long-context tasks well-suited for evaluating compression methods\citep{bai2024longbenchbilingualmultitaskbenchmark}. To ensure fair comparisons, we follow the experimental settings of \citep{shi2024discoveringgemsearlylayers}. Under these settings, we evaluate All KV (the full-attention original model), H2O with a KV cache size of 4096, SnapKV, GemFilter, and PromptDistill, using selected tokens or KV cache sizes of 1024, 2048, and 4096. We conduct preliminary tests on the baseline methods and find that our results closely match those reported in the GemFilter paper. Therefore, we directly use their reported results for baselines and integrate our findings into the table.

In \cref{table:longbench}, we present all results, highlighting our differences compared to GemFilter. The results show that while PromptDistill is significantly more efficient than full attention, it achieves performance close to full attention in general and even surpasses it when using the Qwen2 model with 4096 selected tokens. Compared to H2O, which is less efficient than our method, PromptDistill demonstrates a clear performance advantage. Similarly, while SnapKV is also less efficient than our method, our results remain comparable overall. Finally, compared to GemFilter, which requires slightly more computation time but is more memory-efficient on GPUs, PromptDistill consistently achieves better results across all cases.

\subsubsection{InfBench}\label{sec:infbench}

\begin{table}
  \centering
  %\small
  \caption{Evaluation Results for Infbench (PromptDistill's difference from GemFilter is labeled)}
  \label{table:infbench}
  \resizebox{\linewidth}{!}{
  \begin{tabular}{l*{12}{c}}
  \toprule
  \textbf{Method} & \textbf{passkey} & \textbf{N\_string} & \textbf{kv\_retr} & \textbf{Lbook\_sum\_eng} & \textbf{Lbook\_choice\_eng} & \textbf{Lbook\_qa\_eng} & \textbf{Lbook\_qa\_chn} & \textbf{Ldia\_qa\_eng} & \textbf{math\_find} & \textbf{code\_run} & \textbf{code\_debug} & \textbf{Mean}\\
  \midrule
  %&&&&&&\textbf{LLama 3.1 8B Instruct}\\
  \rowcolor[gray]{.9}&\multicolumn{12}{c}{\textbf{{LLama 3.1 8B Instruct}}}\\
  All KV & 81.36 & 81.36 & 66.6 & 18.86 & 67.25 & 14.85 & 12.35 & 15.0 & 33.14 & 1.0 & 1.27 & 35.73\\
  H2O-4096 & 6.78 & 6.1 & 2.0 & 18.86 & 47.16 & 8.79 & 10.32 & 7.0 & 34.0 & 0.0 & 18.53 & 14.5\\
  \midrule
  SnapKV-1024 & 81.36 & 71.36 & 0.4 & 15.45 & 67.69 & 11.72 & 11.89 & 12.0 & 34.0 & 1.0 & 1.27 & 28.01\\
  GemFilter-1024 & 81.36 & 81.36 & 1.8 & 15.66 & 36.68 & 12.08 & 13.11 & 14.0 & 34.29 & 0.0 & 19.29 & 28.15\\
  PromptDistill-1024 & 81.36 & 78.98 & 2.0 & 16.02 & 65.5 & 12.04 & 12.39 & 9.5 & 32.57 & 1.25 & 1.27 & 28.44(+0.29)\\
  \midrule
  SnapKV-2048 & 81.36 & 77.12 & 1.2 & 16.78 & 67.69 & 13.41 & 11.86 & 10.5 & 34.0 & 1.0 & 1.02 & 28.72\\
  GemFilter-2048 & 81.36 & 81.36 & 4.4 & 17.02 & 43.23 & 11.77 & 12.91 & 17.5 & 31.71 & 0.0 & 19.04 & 29.12\\
  PromptDistill-2048 & 81.36 & 79.49 & 4.0 & 16.94 & 66.38 & 14.28 & 12.2 & 13.0 & 32.57 & 1.25 & 1.02 & 29.32(+0.2)\\
  \midrule
  SnapKV-4096 & 81.36 & 78.98 & 3.6 & 16.44 & 67.69 & 14.12 & 12.21 & 15.5 & 34.0 & 0.75 & 1.02 & 29.61\\
  GemFilter-4096 & 81.36 & 81.36 & 13.0 & 18.05 & 44.54 & 11.81 & 13.93 & 11.5 & 35.14 & 0.0 & 14.97 & 29.61\\
  PromptDistill-4096 & 81.36 & 80.34 & 9.2 & 18.26 & 66.81 & 14.71 & 12.48 & 14.5 & 32.86 & 0.25 & 0.76 & 30.14(+0.53)\\
  \midrule
  %&&&&&&\textbf{Phi 3.5 Mini 3.8B Instruct}\\
  \rowcolor[gray]{.9}&\multicolumn{12}{c}{\textbf{{Phi 3.5 Mini 3.8B Instruct}}}\\
  All KV & 40.68 & 40.68 & 26.8 & 17.68 & 48.03 & 9.1 & 12.88 & 10.0 & 24.57 & 1.0 & 19.29 & 22.79\\
  H2O-4096 & 6.78 & 6.78 & 0.4 & 16.71 & 35.37 & 6.84 & 9.78 & 6.5 & 29.14 & 0.75 & 5.58 & 11.33\\
  \midrule
  SnapKV-1024 & 62.54 & 15.25 & 0.2 & 13.87 & 56.33 & 8.62 & 12.71 & 11.5 & 14.29 & 0.0 & 17.51 & 19.35\\
  GemFilter-1024 & 40.68 & 40.68 & 0.6 & 13.53 & 45.41 & 7.01 & 11.48 & 7.0 & 15.14 & 1.5 & 10.41 & 17.59\\
  PromptDistill-1024 & 40.68 & 40.68 & 1.0 & 15.18 & 46.72 & 7.92 & 12.88 & 11.5 & 22.29 & 1.0 & 17.01 & 19.71(+2.12)\\
  \midrule
  SnapKV-2048 & 40.51 & 15.93 & 0.2 & 15.18 & 44.54 & 8.19 & 12.95 & 12.5 & 26.86 & 1.25 & 20.05 & 18.01\\
  GemFilter-2048 & 40.68 & 40.51 & 2.0 & 14.42 & 45.85 & 7.33 & 11.69 & 7.5 & 19.43 & 2.0 & 16.75 & 18.92\\
  PromptDistill-2048 & 40.68 & 40.68 & 2.0 & 16.0 & 45.85 & 8.13 & 12.53 & 9.0 & 23.71 & 2.0 & 18.02 & 19.87(+0.95)\\
  \midrule
  SnapKV-4096 & 40.68 & 17.29 & 1.4 & 15.83 & 44.54 & 8.33 & 13.2 & 12.0 & 26.29 & 1.75 & 20.05 & 18.31\\
  GemFilter-4096 & 0.0 & 0.0 & 0.0 & 1.53 & 27.51 & 5.17 & 1.38 & 0.5 & 1.71 & 0.0 & 0.0 & 3.44\\
  PromptDistill-4096 & 40.68 & 40.68 & 6.0 & 16.17 & 46.72 & 8.25 & 12.7 & 10.0 & 24.57 & 1.75 & 18.02 & 20.5(+17.06)\\
  \midrule
  %&&&&&&\textbf{Qwen2 7B Instruct}\\
  \rowcolor[gray]{.9}&\multicolumn{12}{c}{\textbf{{Qwen2 7B Instruct}}}\\
  All KV & 27.12 & 27.12 & 13.8 & 16.42 & 54.59 & 6.23 & 10.0 & 7.0 & 28.57 & 0.0 & 26.14 & 19.73\\
  H2O-4096 & 0.85 & 3.05 & 0.0 & 0.0 & 47.6 & 1.41 & 2.11 & 1.0 & 24.29 & 0.0 & 0.0 & 7.3\\
  \midrule
  SnapKV-1024 & 24.75 & 5.25 & 0.4 & 13.92 & 55.02 & 5.91 & 9.93 & 8.0 & 29.14 & 0.0 & 26.9 & 16.29\\
  GemFilter-1024 & 27.12 & 27.12 & 0.2 & 13.43 & 43.67 & 8.17 & 18.22 & 12.0 & 10.0 & 0.0 & 28.43 & 17.12\\
  PromptDistill-1024 & 27.12 & 27.12 & 0.2 & 13.72 & 54.59 & 5.66 & 9.67 & 11.5 & 27.14 & 0.0 & 28.93 & 18.7(+1.58)\\
  \midrule
  SnapKV-2048 & 25.25 & 6.78 & 0.6 & 14.51 & 55.02 & 6.11 & 9.93 & 8.5 & 28.29 & 0.25 & 26.65 & 16.54\\
  GemFilter-2048 & 27.12 & 27.12 & 0.2 & 14.45 & 44.54 & 9.17 & 18.01 & 14.0 & 10.86 & 0.0 & 28.17 & 17.6\\
  PromptDistill-2048 & 27.12 & 27.12 & 0.2 & 15.01 & 55.9 & 5.97 & 9.61 & 12.0 & 28.29 & 0.0 & 28.43 & 19.06(+1.46)\\
  \midrule
  SnapKV-4096 & 26.44 & 9.66 & 1.2 & 15.18 & 55.02 & 6.21 & 9.91 & 9.0 & 29.14 & 0.25 & 26.9 & 17.17\\
  GemFilter-4096 & 27.12 & 27.12 & 0.8 & 15.4 & 44.54 & 7.56 & 12.81 & 16.0 & 12.0 & 0.0 & 28.68 & 17.46\\
  PromptDistill-4096 & 27.12 & 27.12 & 0.6 & 15.74 & 56.33 & 6.09 & 9.59 & 11.0 & 28.29 & 0.0 & 28.68 & 19.14(+1.68)\\
  \bottomrule
  \end{tabular}
  }
\end{table}
For InfBench, a benchmark consisting of long-context tasks with even longer prompts (average data length exceeding 100K tokens), including both synthetic and real-world tasks across diverse domains\citep{zhang2024inftybenchextendinglongcontext}, we adopt the same method settings as in LongBench, although baseline results for InfBench from GemFilter’s report are not available. The results are presented in \cref{table:infbench}. Compared to H2O and GemFilter, PromptDistill once again demonstrates a clear performance advantage, as observed in LongBench evaluations. More importantly, this time PromptDistill achieves a complete performance advantage over SnapKV. The extremely poor performance of the Phi-3.5 model with GemFilter at 4096 selected tokens is explained in \cref{appendix: extreme}.

\begin{wraptable}{r}{0.75\textwidth}
  \centering
  \small
  \caption{Evaluation Results for Needle in a Haystack (PromptDistill's difference from GemFilter is labeled). Instruct versions of the models are used.}
  \label{table:needle}
  \resizebox{0.9\linewidth}{!}{
  \begin{tabular}{lccc}
  \toprule
  \textbf{Method} & \textbf{LLama 3.1 8B} & \textbf{Phi 3.5 Mini 3.8B} & \textbf{Qwen2 7B}\\
  \midrule
  All KV & 85.2 & 91.7 & 91.2\\
  \midrule
  SnapKV-1024 & 64.0 & 91.7 & 90.5\\
  GemFilter-1024 & 81.8 & 91.7 & 69.7\\
  PromptDistill-1024 & 82.6(+0.8) & 91.7(+0) & 71.6(+1.9)\\
  \midrule
  SnapKV-2048 & 71.7 & 91.7 & 90.8\\
  GemFilter-2048 & 84.7 & 91.7 & 81.8\\
  PromptDistill-2048 & 84.3(-0.4) & 91.7(+0) & 82.2(+0.4)\\
  \midrule
  SnapKV-4096 & 77.1 & 91.7 & 91.0\\
  GemFilter-4096 & 83.3 & 5.1 & 89.3\\
  PromptDistill-4096 & 84.6(+1.3) & 91.7(+0) & 90.2(+0.9)\\
  \bottomrule
  \end{tabular}
  }
\end{wraptable}

\subsubsection{Needle in a Haystack}\label{sec:needle}

We also evaluate the methods on Needle in a Haystack, which tests an LLM's ability to retrieve accurate information from a specific sentence hidden within a large document \citep{nelson2024needlehaystackmemorybased}. While GemFilter \citep{shi2024discoveringgemsearlylayers} only evaluates selected tokens or KV cache sizes of 1024, we conduct a more comprehensive evaluation using settings similar to those in LongBench and InfBench.

Since our results differ from those in GemFilter’s report—likely due to differences in hardware, Python, or CUDA versions—we do not use their results as baselines for fair comparison.

\begin{wraptable}{r}{0.6\linewidth}
\vspace{-2pt}
  \centering
  \small
    \caption{Top-5 multi-stage selection performance on LongBench (LLama 3.1 8B Instruct), with $k$ and $r$ denoting the token numbers and stages for selection, respectively (space-separated for multiple stages). Here, $\Delta_1$ is the difference in mean score from PromptDistill-13-1024, and $\Delta_2$ from GemFilter-13-1024.}
  \begin{tabular}{lcccc}
    \toprule
    $\mathbf{k}$ & $\mathbf{r}$ & \textbf{Mean Score} & $\Delta_1$ & $\Delta_2$ \\
    \midrule
    1024 & 13 & 35.36 & 0 & 0.86 \\
%    1024 & 13 & 34.50 & -0.86 & 0 \\
    \midrule
    8192 1024 & 8 13 & 35.39 & 0.03 & 0.89 \\
    16384 8192 1024 & 7 8 13 & 35.42 & 0.06 & 0.92 \\
    16384 8192 1024 & 5 8 13 & 35.53 & 0.17 & 1.03 \\
    16384 8192 1024 & 3 8 13 & 35.42 & 0.06 & 0.92 \\
    16384 8192 1024 & 1 8 13 & 35.32 & -0.04 & 0.82 \\
    \bottomrule
  \end{tabular}
  \label{table:briefmultistage}
\end{wraptable}

For maximum input length, we set 120K for LLaMA 3.1 8B Instruct (following GemFilter's work \citep{shi2024discoveringgemsearlylayers}), 86K for Phi-3.5 Mini 3.8B Instruct (due to GPU memory limitations for the full attention method), and 32.2K for Qwen2 7B Instruct (due to its input length limit). The results are presented in \cref{table:needle}, with additional detailed figures in \cref{appendix: figures for needle}. For Phi-3.5, all method settings yield the same results except GemFilter-4096, which encounters the same issue discussed in \cref{sec:infbench} and in \cref{appendix: extreme}. Overall, PromptDistill outperforms GemFilter in most cases and surpasses SnapKV in half of the evaluations.

\subsection{Analysis}\label{sec:analysis}
We empirically determine the selection layer for Qwen2 7B Instruct in our main evaluations by assessing GemFilter across different layers on LongBench. We select the layer where GemFilter performs well while ensuring it is as early as possible for efficiency. This approach ensures a fair comparison, as choosing a layer where GemFilter performs well highlights its selection ability. Further details on this selection process are provided in \cref{appendix: layer choice experiments}. In the remainder of this section, we present an ablation study on different settings of PromptDistill.
\subsubsection{Multi-stage Selection Analysis}\label{sec:multi-stage selection analysis}
% \begin{table}
%   \centering
%   \small
%   \caption{Top-5 Multi-stage selection performance on LongBench (LLama 3.1 8B Instruct)}
%   \label{table:briefmultistage}
%   \resizebox{0.8\linewidth}{!}{
%   \begin{tabular}{lcccc}
%   \toprule
%   \textbf{selection token nums} & \textbf{selection stages} & \textbf{Mean Score} & \textbf{delta from PromptDistill-13-1024} & \textbf{delta from GemFilter-13-1024} \\
%   \midrule
%   1024 & 13 & 35.36 & 0 & 0.86\\
%   1024 & 13 & 34.50 & -0.86 & 0\\
%   \midrule
%   8192 1024 & 8 13 & 35.39 & 0.03 & 0.89\\
%   16384 8192 1024 & 7 8 13 & 35.42 & 0.06 & 0.92\\
%   16384 8192 1024 & 5 8 13 & 35.53 & 0.17 & 1.03\\
%   16384 8192 1024 & 3 8 13 & 35.42 & 0.06 & 0.92\\
%   16384 8192 1024 & 1 8 13 & 35.32 & 0.04 & 0.82\\
%   \bottomrule
%   \end{tabular}
%   }
% \end{table}

For PromptDistill with multi-stage selection, various configurations can be chosen. To explore its effectiveness, we design a simple evaluation method using LongBench. As a baseline, we apply PromptDistill with a single selection of 1024 tokens at layer 13, and for all trials, the KV cache is truncated to 1024 tokens. We present a set of well-performing configurations in \cref{appendix: multi} (\cref{table:multistage}), and display a brief subset with the highest performance here in \cref{table:briefmultistage}.
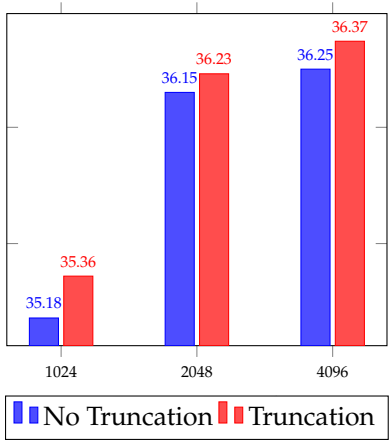
\begin{wrapfigure}{r}{0.5\textwidth}
\vspace{-2pt}
  \centering
  \begin{tikzpicture}
    \begin{axis}[
      ybar,
      bar width=11pt,
      width=0.95\linewidth,  % Adjusted to fit within the wrapfigure
      height=6cm,
      enlarge x limits=0.2,
      xlabel={Context Length},
      symbolic x coords={1024,2048,4096},
      xtick=data,
      xticklabel style={font=\tiny},
      yticklabels={},
      nodes near coords,
      nodes near coords style={font=\tiny},
      legend style={at={(0.5,-0.15)}, anchor=north, legend columns=-1},
    ]
      \addplot+[fill=blue!70] coordinates {(1024,35.18) (2048,36.15) (4096,36.25)};
      \addplot+[fill=red!70] coordinates {(1024,35.36) (2048,36.23) (4096,36.37)};
      \legend{No Truncation, Truncation}
    \end{axis}
  \end{tikzpicture}
  \caption{The effect of cache truncation.}
  \label{fig:cache-truncation}
  \vspace{-2pt}
\end{wrapfigure}

Our exploration begins by introducing an additional selection layer that selects 2048 tokens before layer 13. If certain layers yield good performance, we add them to the baseline set and continue exploring layers capable of selecting 4096 tokens. We find that selecting 2048 tokens in earlier layers leads to noticeable performance degradation, so we focus on identifying layers for 4096 tokens. Layer 10 performs well in this regard, prompting us to test 8192 tokens—first on top of 1024 tokens at layer 13, then on top of 4096 tokens at layer 10, followed by 1024 tokens at layer 13. This process yields four optimal configurations, shown in the third to sixth rows of \cref{table:multistage}. Further exploration of 16384-token selection layers provides the remaining results. Both the two tables clearly highlight differences between each setting and GemFilter/PromptDistill with single-stage selection. Some multi-stage configurations achieve comparable performance to single-stage PromptDistill, while others trade off some accuracy for higher efficiency yet still outperform GemFilter. The theoretical analysis in \cref{sec:theoretical analysis} demonstrates the efficiency gains of multi-stage selection. Here, empirical results further confirm that many configurations can maintain performance, indicating that multi-stage selection has strong potential to reduce the memory efficiency gap between PromptDistill and GemFilter while significantly improving time efficiency over GemFilter, all while preserving performance advantages.

\subsubsection{Cache Truncation Analysis}\label{sec:cache truncation analysis}

% \begin{table}
%   \centering
%   %\small
%   \caption{Comparison for Cache Truncation on LongBench (LLama 3.1 8B Instruct)}
%   \label{table:truncation}
%   \resizebox{\linewidth}{!}{
%   \begin{tabular}{l*{15}{c}}
%   \toprule
%   \textbf{Method} & \textbf{NrtvQA} & \textbf{qasper} & \textbf{MF-en} & \textbf{hotpotqa} & \textbf{2wikimqa} & \textbf{musique} & \textbf{gov\_report} & \textbf{qmsum} & \textbf{multi\_news} & \textbf{trec} & \textbf{triviaqa} & \textbf{samsum} & \textbf{PCount} & \textbf{PRe} & \textbf{Mean}\\
%   \midrule
%   PromptDistill-1024 (no truncation) & 23.1 & 10.87 & 26.61 & 16.12 & 15.47 & 11.67 & 29.88 & 22.86 & 25.93 & 73.5 & 91.57 & 43.62 & 5.41 & 95.97 & 35.18\\
%   PromptDistill-1024 & 24.24 & 11.36 & 28.01 & 16.59 & 14.92 & 11.7 & 29.68 & 22.29 & 25.77 & 73.5 & 91.65 & 43.61 & 5.87 & 95.8 & 35.36\\
%   \midrule
%   PromptDistill-2048 (no truncation) & 27.62 & 12.61 & 26.87 & 16.51 & 16.39 & 12.57 & 31.99 & 23.39 & 26.57 & 73.5 & 91.25 & 43.92 & 5.26 & 97.67 & 36.15\\
%   PromptDistill-2048 & 28.24 & 12.86 & 26.75 & 17.47 & 15.88 & 12.46 & 32.06 & 23.0 & 26.86 & 73.5 & 91.54 & 43.22 & 5.62 & 97.79 & 36.23\\
%   \midrule
%   PromptDistill-4096 (no truncation) & 28.63 & 13.2 & 26.98 & 16.23 & 16.35 & 11.45 & 33.67 & 23.41 & 26.82 & 72.5 & 91.31 & 44.11 & 5.02 & 97.82 & 36.25\\
%   PromptDistill-4096 & 30.2 & 13.24 & 26.91 & 16.41 & 16.12 & 11.6 & 34.2 & 23.08 & 26.86 & 72.5 & 91.47 & 44.17 & 5.24 & 97.22 & 36.37\\
%   \bottomrule
%   \end{tabular}
%   }
% \end{table}
Here, we present \cref{fig:cache-truncation}, a bar chart comparing PromptDistill with and without cache truncation on LongBench using Llama 3.1 8B Instruct (selection layer 13, selected token counts 1024, 2048, 4096). The results clearly show that performance does not decrease and even improves in these cases. Since PromptDistill with cache truncation is significantly more efficient than the basic design without truncation (discussed in \cref{sec:theoretical analysis}) while maintaining performance, we confidently adopt cache truncation for both higher efficiency and stable performance.
\section{Conclusion}
PromptDistill is a novel, training-free approach to improving the inference efficiency of large language models while maintaining output quality. By selecting and retaining the hidden states of the most informative tokens in intermediate layers, our method achieves effective compression without discarding critical contextual information. Compared to prior works like GemFilter, SnapKV, and H2O, PromptDistill balances efficiency and performance more effectively, reducing computation and memory costs with minimal degradation. Additionally, our exploration of multi-stage selection further enhances efficiency while preserving model effectiveness. Through extensive experiments across multiple LLMs and diverse evaluation datasets, we demonstrate the superiority and generalizability of PromptDistill, highlighting its effectiveness in optimizing LLM inference for long-context scenarios.

\bibliography{colm2025_conference}
\bibliographystyle{colm2025_conference}

\appendix
\section{Appendix}
\subsection{Related Work}

\subsubsection{Acceleration of Long-Context LLM Inference}
As large language models (LLMs) are increasingly applied to long-context tasks, the computational and memory overhead during inference has become a critical bottleneck. FastGen \citep{ge2023model} adaptively selects one of five KV cache compression methods based on attention head profiling, reducing generation overhead. StreamingLLM \citep{xiao2023efficient} mitigates the “attention sink” issue by introducing placeholder tokens and a local attention mechanism. MInference \citep{jiang2024minference10acceleratingprefilling} leverages sparse computation patterns like A-shape, Vertical-Slash, and Block-Sparse to reduce latency in the prefilling stage. ThinK \citep{xu2025thinkthinnerkeycache} prunes redundant attention channels by observing imbalances in the key cache's channel dimension. SnapKV \citep{li2024snapkvllmknowslooking} clusters KV positions to focus on key prompt features. Mistral \citep{jiang2023mistral7b} employs a sliding window strategy for efficient long-text inference, while KIVI \citep{liu2024kivi} applies 2-bit KV cache quantization to reduce memory and computation costs. LongGen \citep{ge2024littlegoeslongway} combines sparse attention patterns with lightweight training strategies, while RefreshKV \citep{xu2025refreshkvupdatingsmallkv} dynamically switches between full-context and partial token attention to improve efficiency. However, our work builds upon these strategies by focusing on identifying informative prompt tokens in intermediate layers. This approach enables efficient token selection without additional training, achieving minimal performance degradation. Furthermore, by leveraging this structure during tuning, we demonstrate improved adaptability of the model under extreme compression scenarios.

\subsubsection{Prompt Selection for Computation Reduction}
Minimizing the number of prompt tokens processed during computation is a straightforward yet powerful method for enhancing efficiency. QLLM \citep{li2024quickllamaqueryawareinferenceacceleration} partitions text into blocks and selects relevant ones using query tokens. H2O \citep{zhang2023h2oheavyhitteroracleefficient} retains only influential tokens, while Scissorhands \citep{liu2023scissorhandsexploitingpersistenceimportance} discards less relevant tokens based on the “importance persistence hypothesis.” Retrieval-augmented speculative Decoding \citep{chen2025longcontextinferenceretrievalaugmentedspeculative} incorporates a retrieval-augmented generation (RAG) strategy, where a draft LLM generates content based on a simplified context. RetrievalAttention \citep{liu2024retrievalattentionacceleratinglongcontextllm} indexes and searches KV cache vectors to selectively retain only information relevant to the current generation. RECOMP \citep{xu2023recompimprovingretrievalaugmentedlms} integrates extraction and summarization techniques to preprocess the original context before feeding it as a prompt, effectively condensing the information. Deja Vu \citep{pmlr-v202-liu23am} trains a lightweight algorithm to predict context sparsity at each layer, providing guidance for token selection in subsequent steps. Selective Context \citep{li2023compressingcontextenhanceinference} leverages self-information calculations to prune redundant lexical units directly. Meanwhile, Selective Attention \citep{NEURIPS2024_14fc4a68} introduces targeted modulation within the self-attention mechanism, allowing the model to adjust its attention distribution dynamically based on context importance. ATTRIEVAL \citep{zhang2025attentionrevealstokenstrainingfree} employs attention mechanisms to better capture implicit facts, ensuring minimal information loss while reducing token counts. Unlike these approaches, which mainly reduce input context tokens, our method targets effective token selection across intermediate layers, achieving greater resource efficiency while maintaining strong performance in constrained environments.

\subsection{Case of extreme performance for GemFilter}\label{appendix: extreme}
We observe that in InfBench and Needle in a Haystack, the Phi-3.5 model with GemFilter performs exceptionally poorly when the selected token count is 4096. After further investigation, we find that on our machines, Phi-3.5 fails to generate meaningful responses when the input length is exactly 4096, whereas inputs of 4000 or 4100 tokens still yield high-quality outputs. This issue may stem from factors such as the specific hardware used, CUDA version differences, or the fine-tuning process of the Phi-3.5 model. Since GemFilter selects important tokens and reruns them, the 4096-token prompt length disrupts its performance.

\subsection{More Algorithms}\label{appendix: more algorithms}
Two of the algorithms described in \cref{sec:promptdistill} are presented here.

\begin{algorithm}[tbh]
   \caption{LLM-forward-PromptDistill (single-stage selection, without cache truncation)}
   \label{alg:promptdistill-basic}
    \begin{algorithmic}
        \STATE {\bfseries Input:} LLM $M_{0:R}$, embedded input prompt $X$, selection layer $r$, number of selected tokens $k$
        \STATE $\#$ $M$ has $R+1$ layers in total
        \STATE $\#$ $X$ has token length = $n$
        \STATE Cache $\gets$ [ ]
        \FOR{$i \gets 0 ... R$}
            \STATE $X, Q^{(i)}_n, K^{(i)}, V^{(i)}\gets M_i$($X$)
            \STATE Cache$_i \gets K^{(i)}, V^{(i)}$ 
            \IF{$i$  equals  $r$}
                \STATE indices $\gets$ Select$(Q^{(i)}_n, K^{(i)}, k)$
                \STATE $X \gets X_{\text{indices}}$ 
            \ENDIF
        \ENDFOR
        \STATE return generate($X$), Cache
    \end{algorithmic}
\end{algorithm}

\begin{algorithm}[tbh]
   \caption{LLM-forward-PromptDistill (multi-stage selection, with cache truncation)}
   \label{alg:promptdistill-multi}
    \begin{algorithmic}
        \STATE {\bfseries Input:} LLM $M_{0:R}$, embedded input prompt $X$, selection layers $r$, numbers of selected tokens $k$, truncation times $tt$
        \STATE $\#$ $r,k$ are lists with indices $0:m-1$
        \STATE $\#$ $n$ always refer to the end index of $X$
        \STATE Cache $\gets$ [ ]
        \FOR{$i \gets 0 ... R$}
            \STATE $X, Q^{(i)}_n, K^{(i)}, V^{(i)}\gets M_i$($X$)
            \STATE Cache$_i \gets K^{(i)}, V^{(i)}$ 
            \IF{$i$  equals  $r_p$ for any $p$}
                \STATE indices $\gets$ Select$(Q^{(i)}_n, K^{(i)}, k_p)$
                \STATE $X \gets X_{\text{indices}}$ 
                \IF{$p<tt$}
                    \FOR{$j \gets 0 ... i$}
                        \STATE $K^{(j)}, V^{(j)} \gets $Cache$_j$ 
                        \STATE $K^{(j)} \gets K^{(j)}_{\text{indices}}$
                        \STATE $V^{(j)} \gets V^{(j)}_{\text{indices}}$
                        \STATE Cache$_j \gets K^{(j)}, V^{(j)}$ 
                    \ENDFOR
                \ENDIF
            \ENDIF
        \ENDFOR
        \STATE return generate($X$), Cache
    \end{algorithmic}
\end{algorithm}

\subsection{Multi-stage selection performance}\label{appendix: multi}
\begin{figure*}[ht]
    \begin{center}
        \includegraphics[width=.8\textwidth]{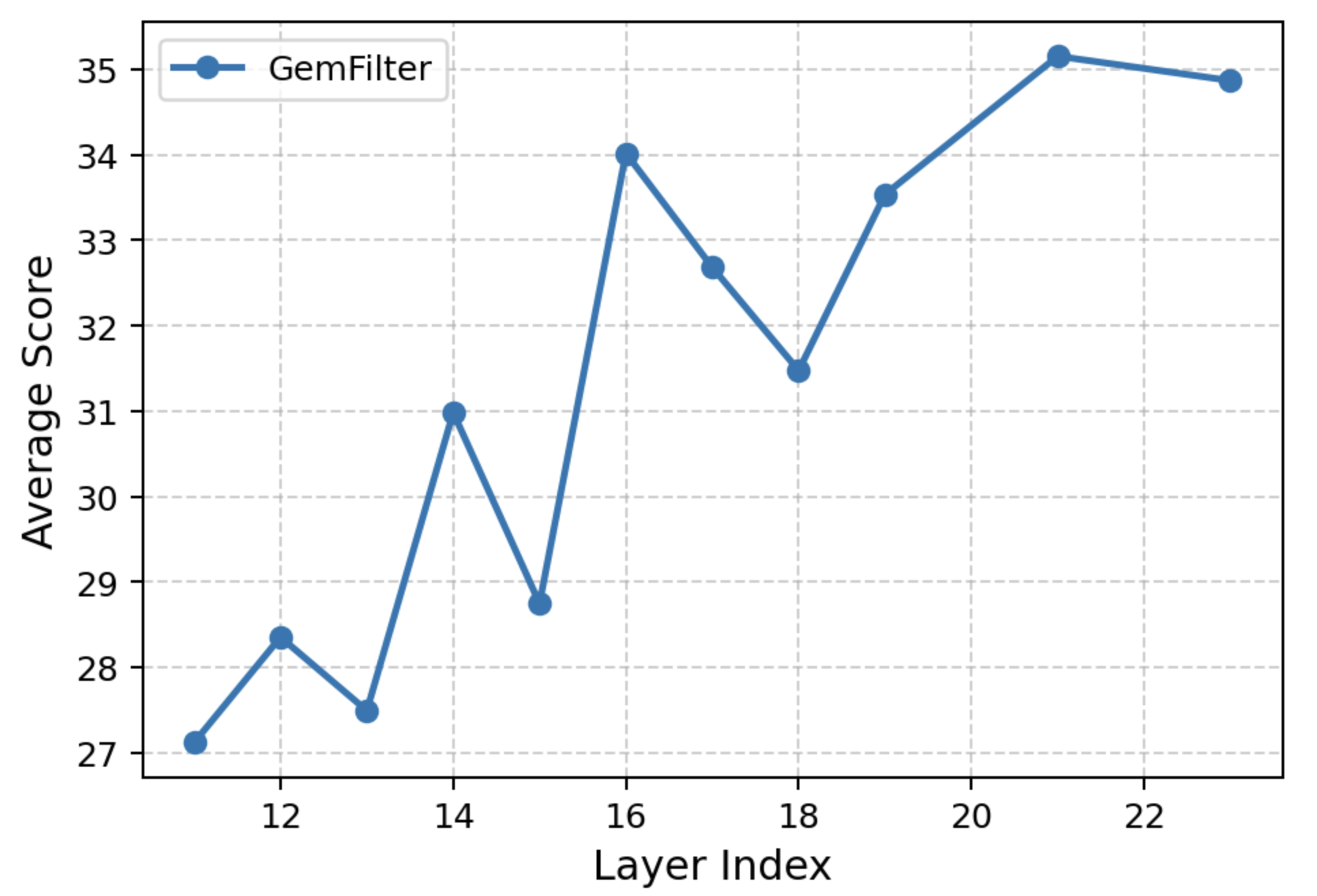}
        \caption{Performance of GemFilter + Qwen2 7B Instruct on LongBench with different selection layers.}
        \label{fig:gem_curve}
    \end{center}
\end{figure*}

Here in \cref{table:multistage}, we present the complete table of Multi-stage Selection performance. For the Qwen2-7B Instruct model, which consists of 28 layers, we need to select a layer where GemFilter performs well - demonstrating a strong ability to identify important tokens — while also ensuring efficiency by avoiding layers that are too late in the sequence. As shown in \cref{fig:gem_curve}, GemFilter achieves a sudden performance boost at layer 16, reaching a score of 34.02, significantly higher than neighboring layers. Although higher scores are observed at layer 21 (35.15) and layer 23 (34.87), these layers are too late in the model, leading to reduced efficiency. Since layer 16 achieves a performance close to the best-performing layers while maintaining better efficiency, we select it as the final selection layer.

\begin{table}
  \centering
  \small
  \caption{Multi-stage selection performance on LongBench (LLama 3.1 8B Instruct)}
  \label{table:multistage}
  \resizebox{\linewidth}{!}{
  \begin{tabular}{lcccc}
  \toprule
  \textbf{Selection token nums} & \textbf{Selection stages} & \textbf{Mean Score} & \textbf{Delta from PromptDistill-13-1024} & \textbf{Delta from GemFilter-13-1024} \\
  \midrule
  1024 & 13 & 35.36 & 0 & 0.86\\
  \midrule
  4096 1024 & 10 13 & 35.00 & -0.36 & 0.5\\
  \midrule
  8192 1024 & 8 13 & 35.39 & 0.03 & 0.89\\
  8192 1024 & 5 13 & 34.78 & -0.58 & 0.28\\
  8192 1024 & 4 13 & 34.67 & -0.69 & 0.17\\
  8192 4096 1024 & 8 10 13 & 35.03 & -0.33 & 0.53\\
  \midrule
  16384 8192 1024 & 7 8 13 & 35.42 & 0.06 & 0.92\\
  16384 8192 1024 & 6 8 13 & 35.38 & 0.02 & 0.88\\
  16384 8192 1024 & 5 8 13 & 35.53 & 0.17 & 1.03\\
  16384 8192 1024 & 4 8 13 & 35.32 & -0.04 & 0.82\\
  16384 8192 1024 & 3 8 13 & 35.42 & 0.06 & 0.92\\
  16384 8192 1024 & 2 8 13 & 35.29 & -0.07 & 0.79\\
  16384 8192 1024 & 1 8 13 & 35.32 & 0.04 & 0.82\\
  16384 8192 1024 & 0 8 13 & 35.30 & -0.06 & 0.8\\
  16384 8192 1024 & 1 5 13 & 34.94 & -0.42 & 0.44\\
  16384 8192 1024 & 0 5 13 & 34.81 & -0.55 & 0.31\\
  16384 8192 1024 & 3 4 13 & 34.81 & -0.55 & 0.31\\
  16384 8192 1024 & 2 4 13 & 34.67 & -0.69 & 0.17\\
  16384 8192 1024 & 1 4 13 & 34.64 & -0.72 & 0.14\\
  16384 8192 1024 & 0 4 13 & 34.57 & -0.79 & 0.07\\
  16384 8192 4096 1024 & 5 8 10 13 & 35.00 & -0.36 & 0.5\\
  16384 8192 4096 1024 & 4 8 10 13 & 34.95 & -0.41 & 0.45\\
  16384 8192 4096 1024 & 3 8 10 13 & 34.92 & -0.44 & 0.42\\
  16384 8192 4096 1024 & 2 8 10 13 & 34.85 & -0.51 & 0.35\\
  16384 8192 4096 1024 & 1 8 10 13 & 34.90 & -0.46 & 0.4\\
  16384 8192 4096 1024 & 0 8 10 13 & 34.80 & -0.56 & 0.3\\
  \bottomrule
  \end{tabular}
  }
\end{table}

\subsection{Layer choice experiments}\label{appendix: layer choice experiments}

\subsection{Figures for Needle in a Haystack}\label{appendix: figures for needle}
Here, we present all figures illustrating the detailed Needle in a Haystack performance, corresponding to the scores shown in \cref{table:needle}. Each graph corresponds to a base model with a specific compression method setting. The grids in the graphs display scores for different context sizes, depth percentages (needle locations within the context).

\begin{figure*}[ht]
    \begin{center}
        \subfigure[LLama 3.1 8B Instruct + All KV: Score = 85.2]{
            \includegraphics[width=.46\textwidth]{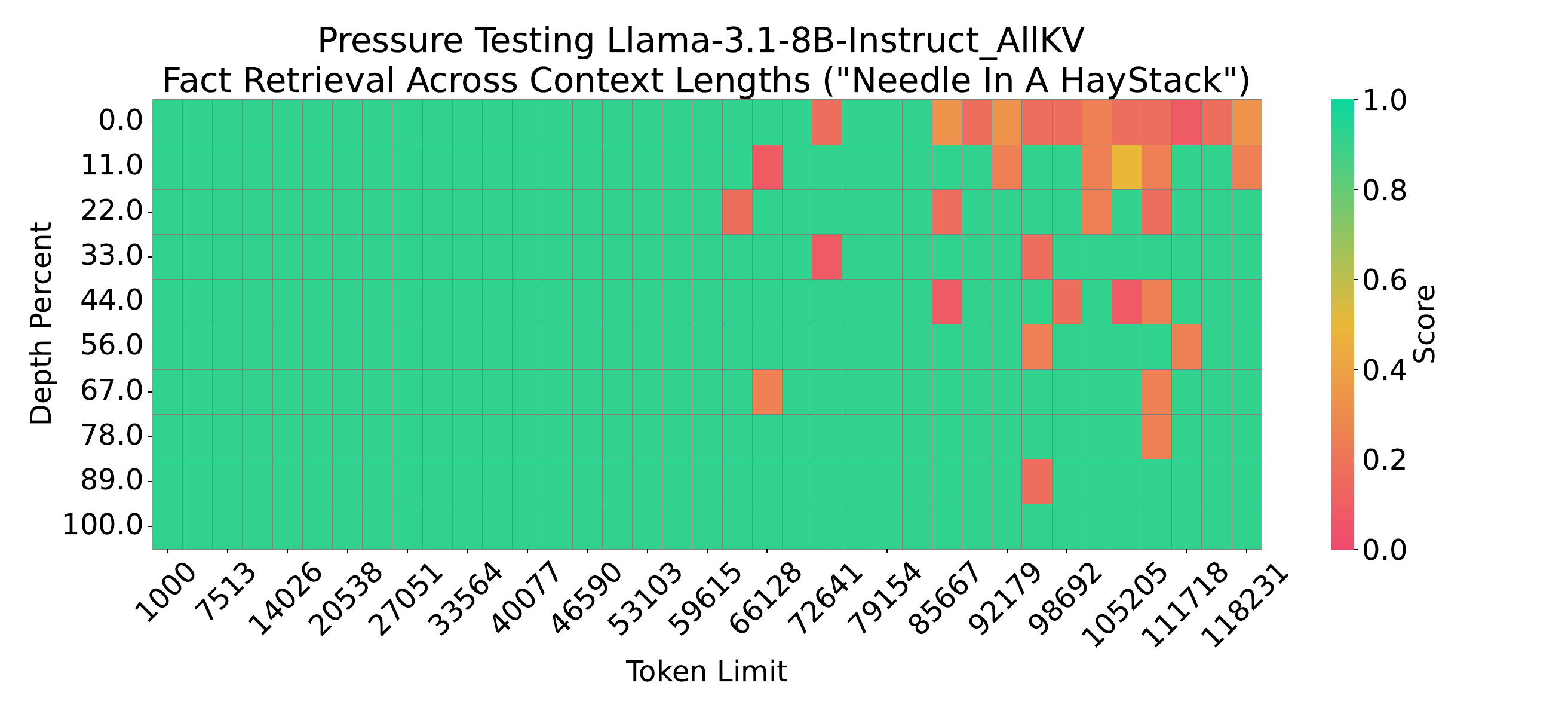}
            \label{fig:llamadefault}
        }
        \subfigure[Qwen2 7B Instruct + All KV: Score = 91.2]{
            \includegraphics[width=.46\textwidth]{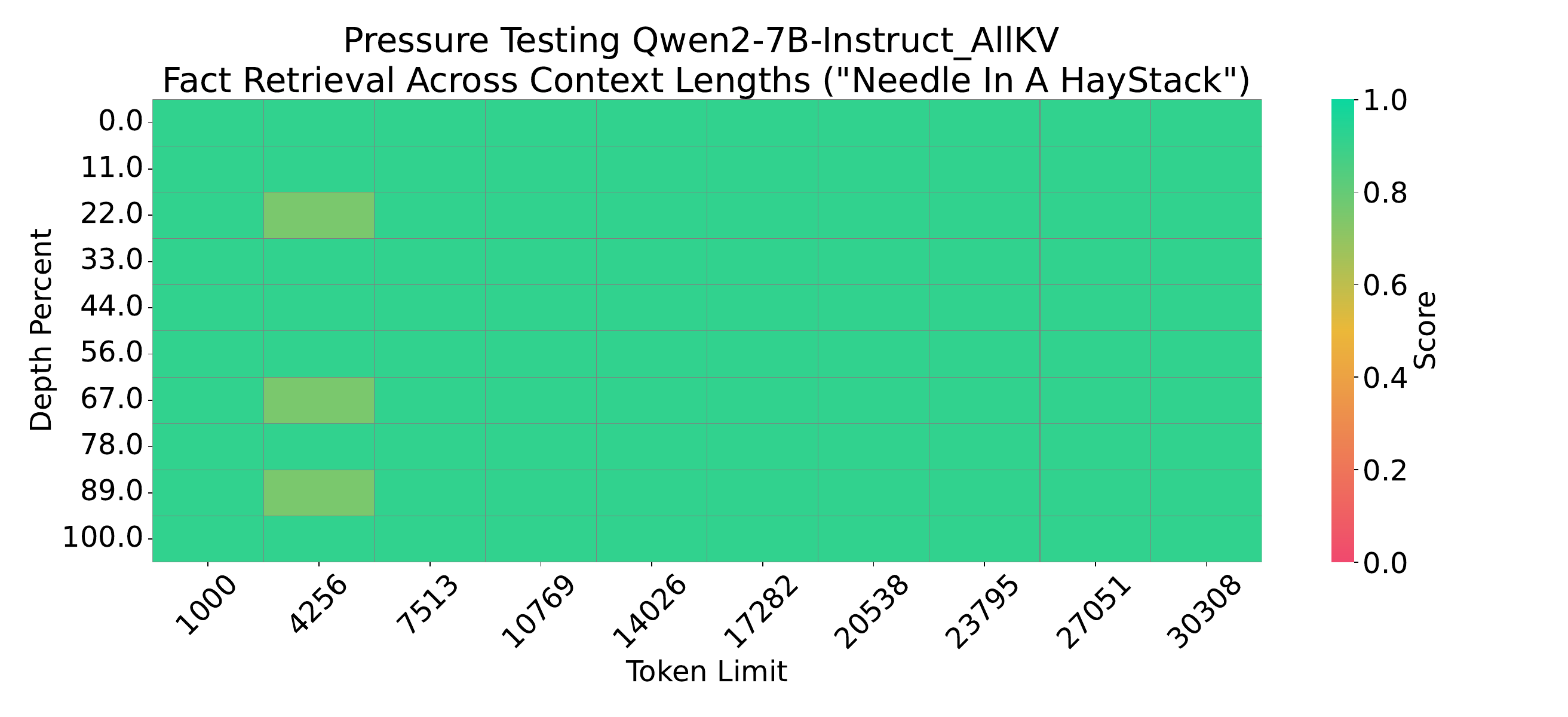}
            \label{fig:qwendefault}
        }
        \subfigure[LLama 3.1 8B Instruct + PromptDistill-1024: Score = 82.6]{
            \includegraphics[width=.46\textwidth]{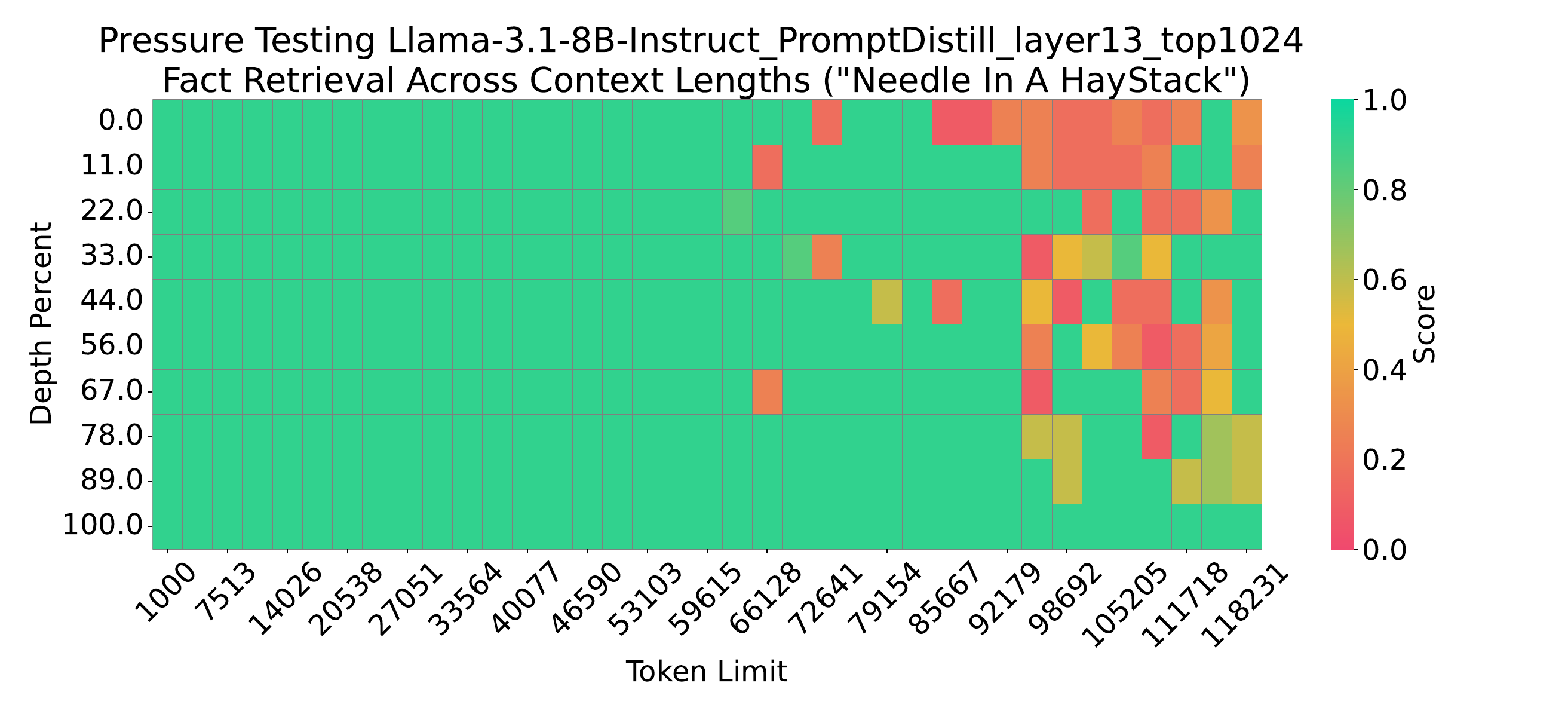}
            \label{fig:llamadist1024}
        }
        \subfigure[Qwen2 7B Instruct + PromptDistill-1024: Score = 71.6]{
            \includegraphics[width=.46\textwidth]{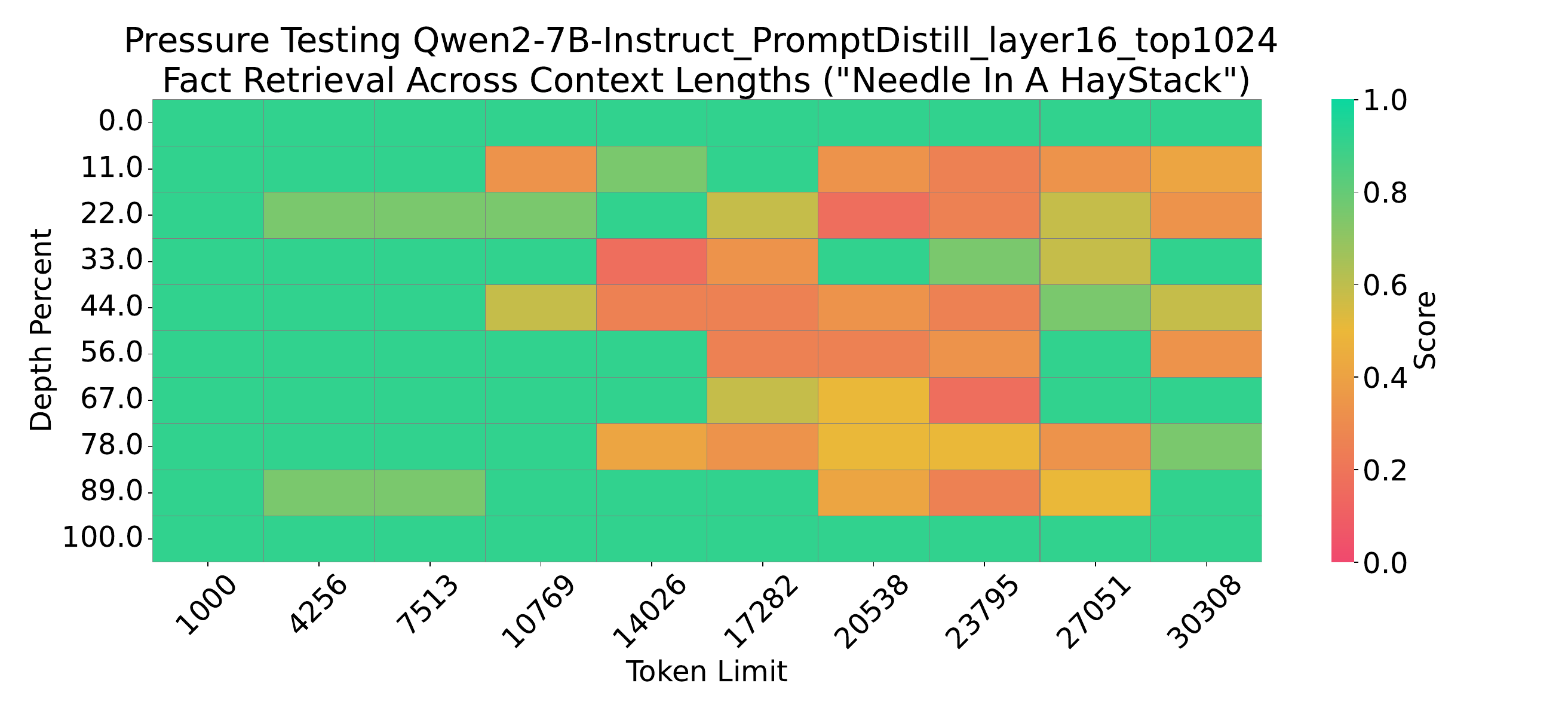}
            \label{fig:qwendist1024}
        }
        \subfigure[LLama 3.1 8B Instruct + GemFilter-1024: Score = 81.8]{
            \includegraphics[width=.46\textwidth]{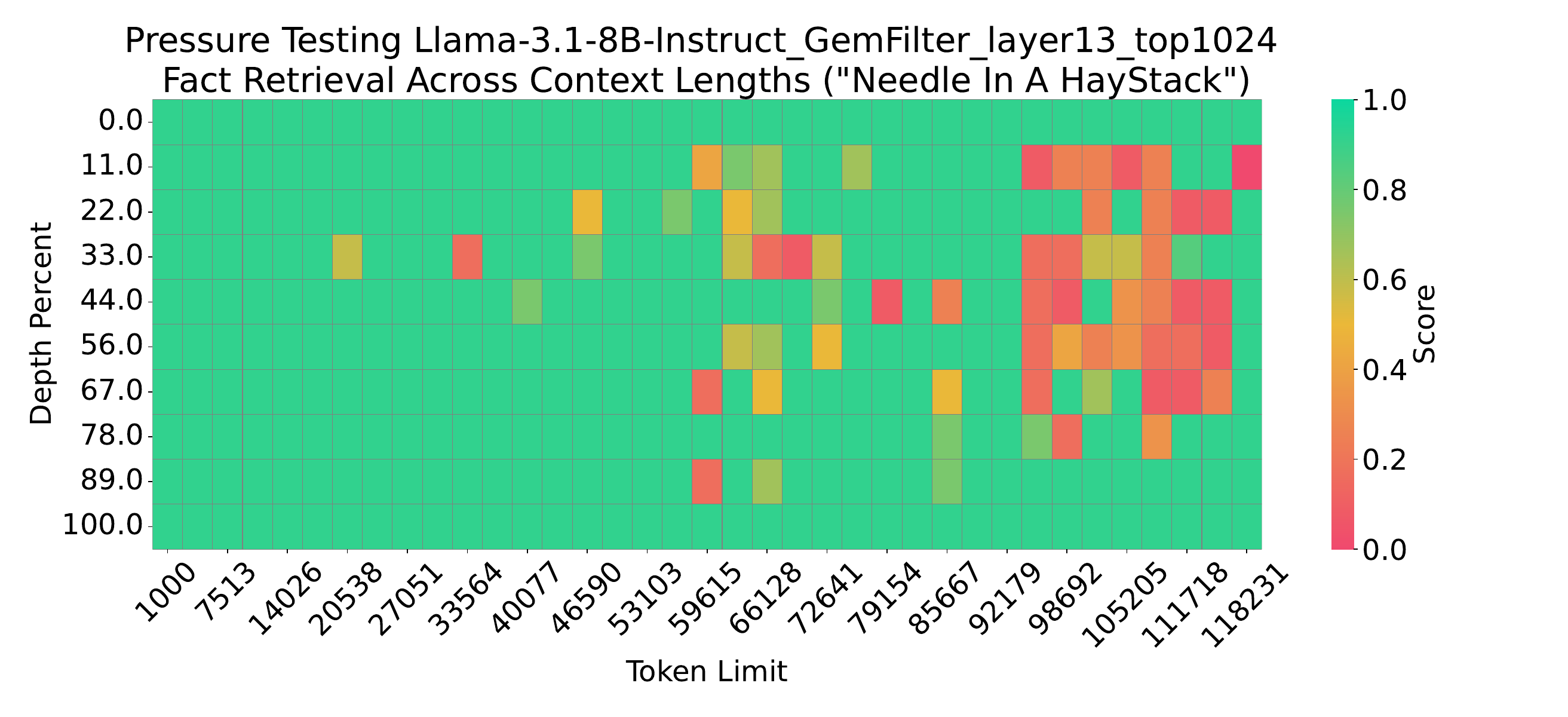}
            \label{fig:llamagem1024}
        }
        \subfigure[Qwen2 7B Instruct + GemFilter-1024: Score = 69.7]{
            \includegraphics[width=.46\textwidth]{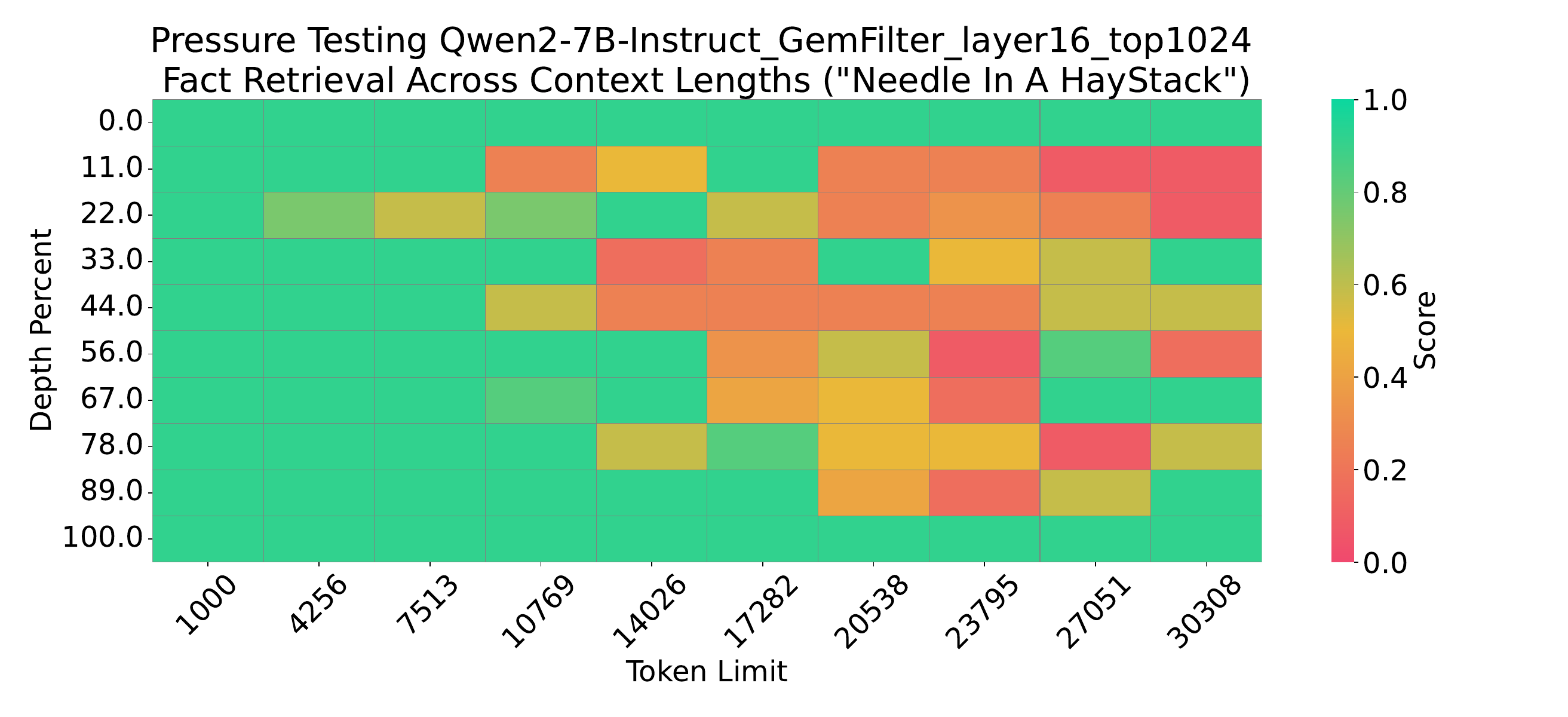}
            \label{fig:qwengem1024}
        }
        \subfigure[LLama 3.1 8B Instruct + SnapKV-1024: Score = 64.0]{
            \includegraphics[width=.46\textwidth]{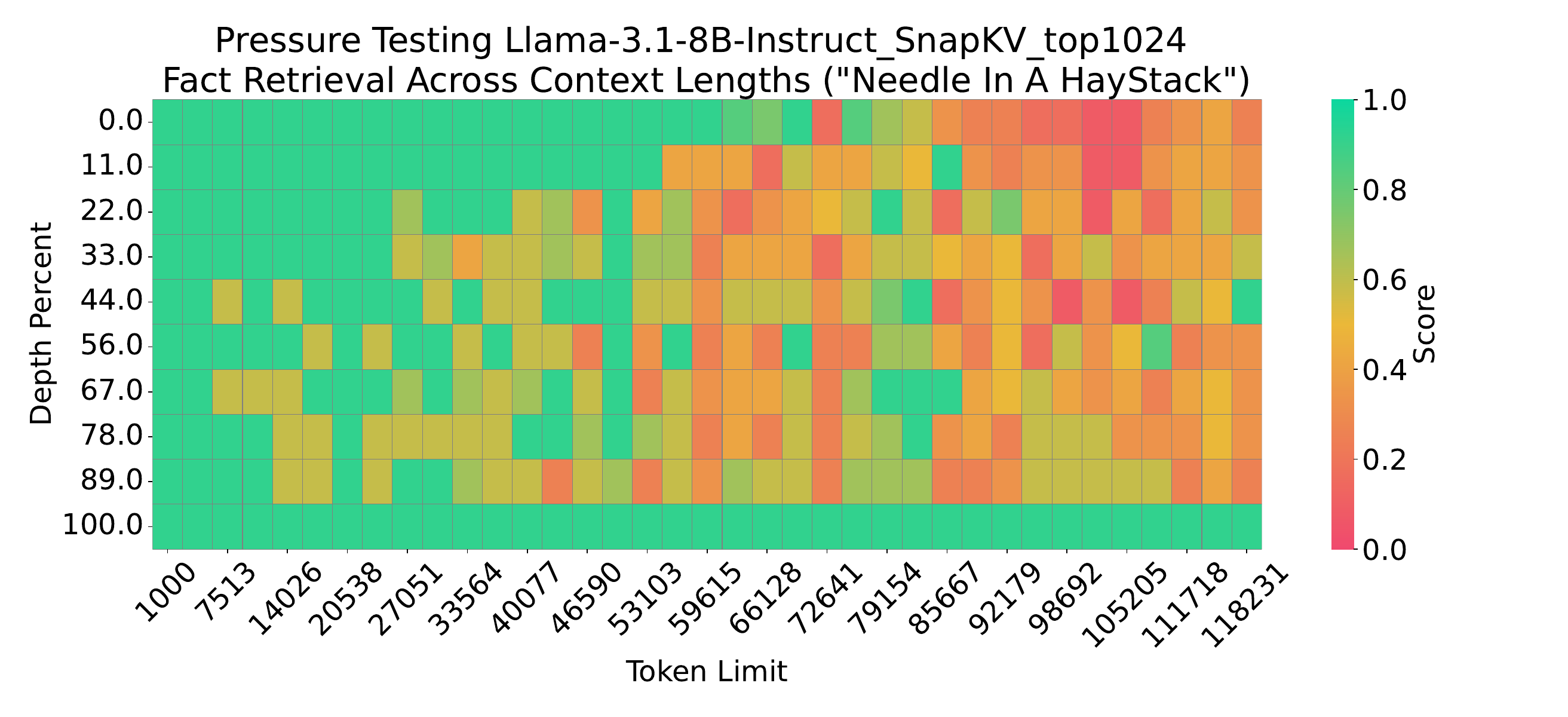}
            \label{fig:llamasnap1024}
        }
        \subfigure[Qwen2 7B Instruct + SnapKV-1024: Score = 90.5]{
            \includegraphics[width=.46\textwidth]{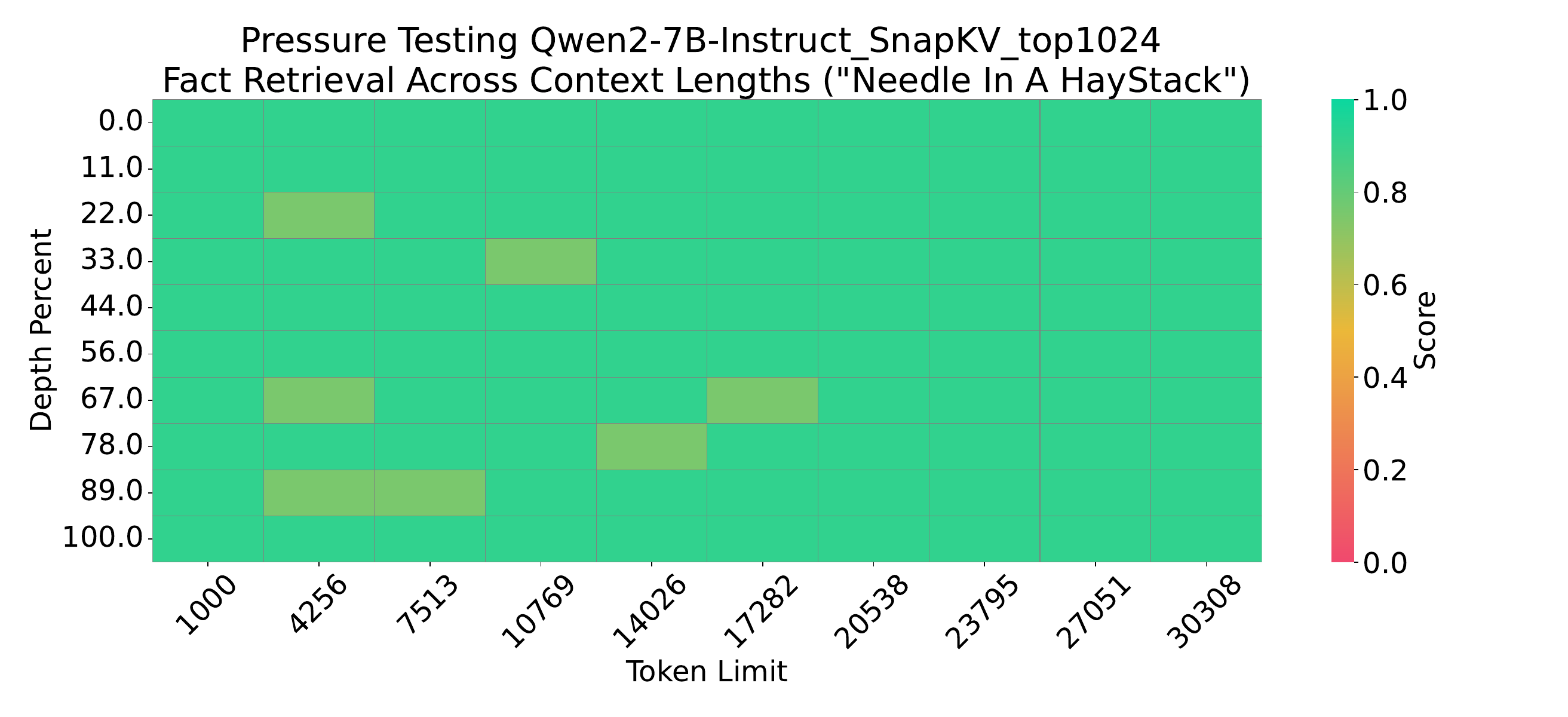}
            \label{fig:qwensnap1024}
        }
        \caption{Evaluation Results for Needle in a Haystack: LLaMA 3.1 8B Instruct's and Qwen2 7B Instruct's scores in all test cases with model settings AllKV, PromptDistill-1024, GemFilter-1024, SnapKV-1024. (The x-axis represents the length of the input tokens, while the y-axis shows the position depth percentage of the ‘needle’ information)}
    \end{center}
    \label{fig:needle1}
\end{figure*}

\begin{figure*}[ht]
    \begin{center}
        \subfigure[LLama 3.1 8B Instruct + PromptDistill-2048: Score = 84.3]{
            \includegraphics[width=.46\textwidth]{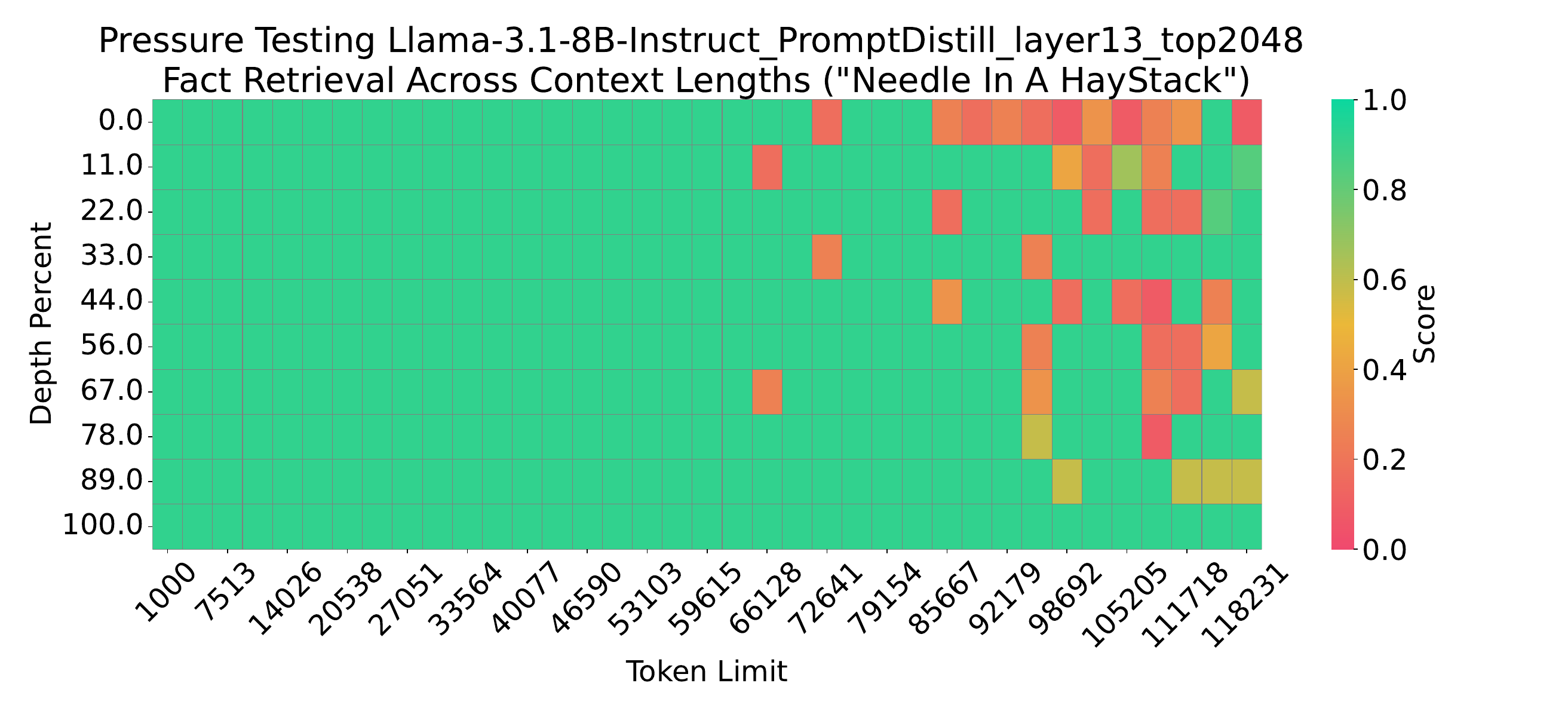}
            \label{fig:llamadist2048}
        }
        \subfigure[Qwen2 7B Instruct + PromptDistill-2048: Score = 82.2]{
            \includegraphics[width=.46\textwidth]{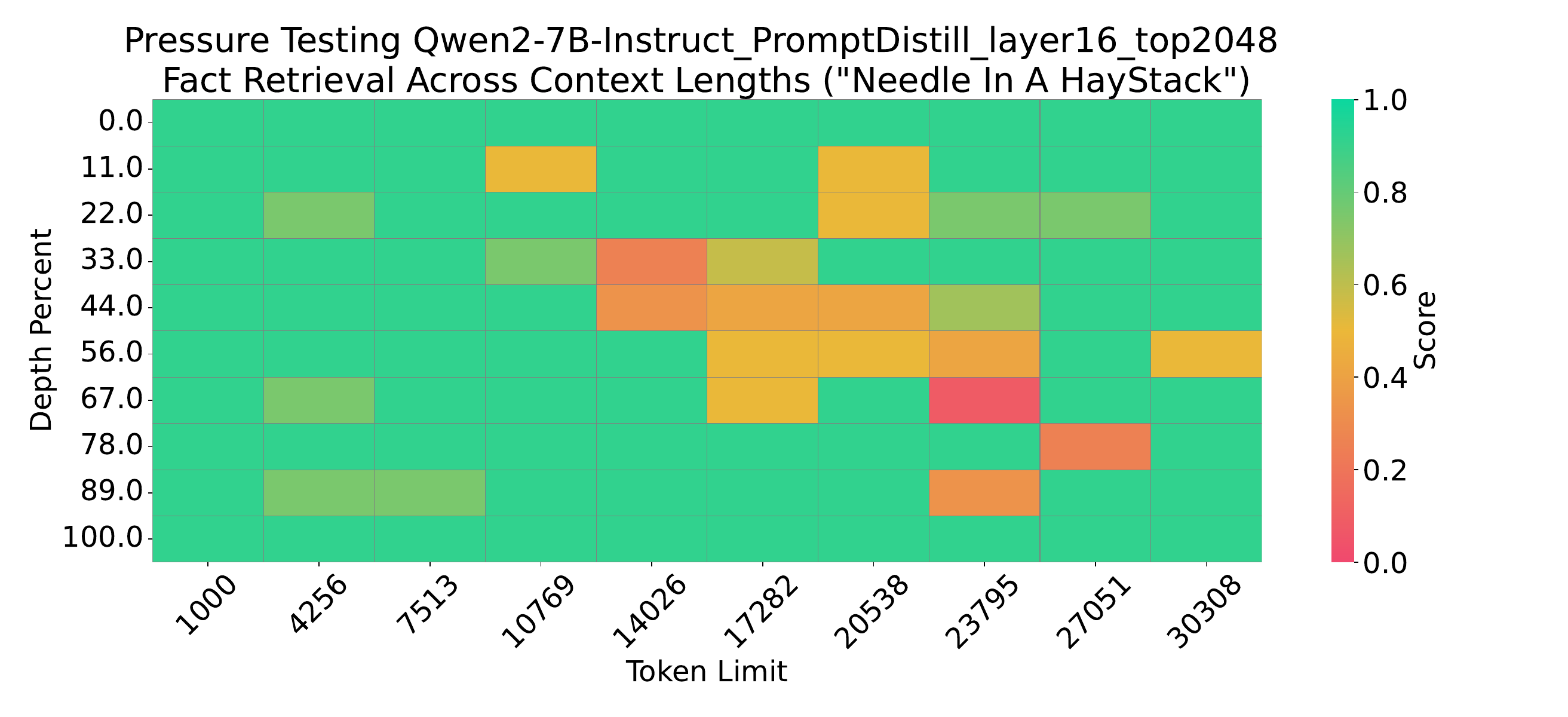}
            \label{fig:qwendist2048}
        }
        \subfigure[LLama 3.1 8B Instruct + GemFilter-2048: Score = 84.7]{
            \includegraphics[width=.46\textwidth]{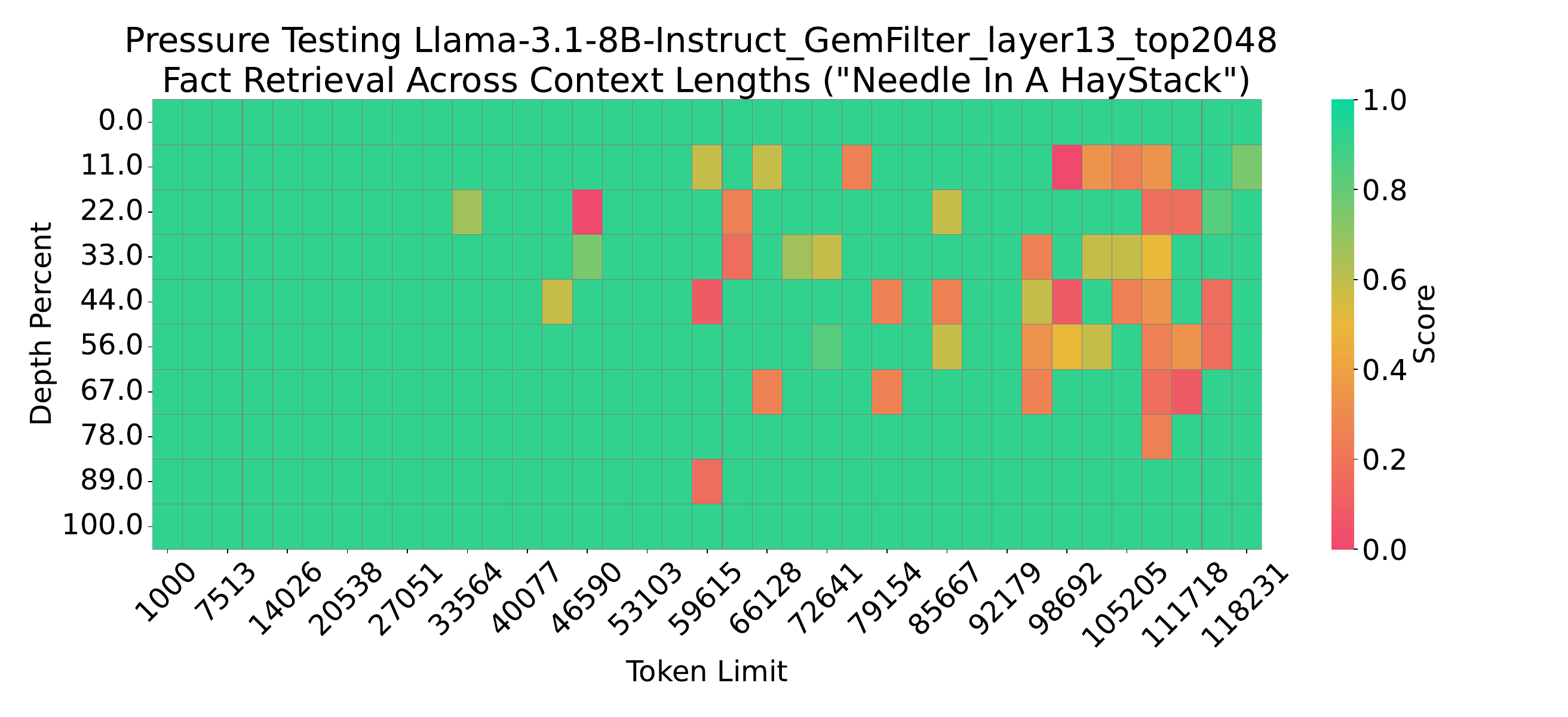}
            \label{fig:llamagem2048}
        }
        \subfigure[Qwen2 7B Instruct + GemFilter-2048: Score = 81.8]{
            \includegraphics[width=.46\textwidth]{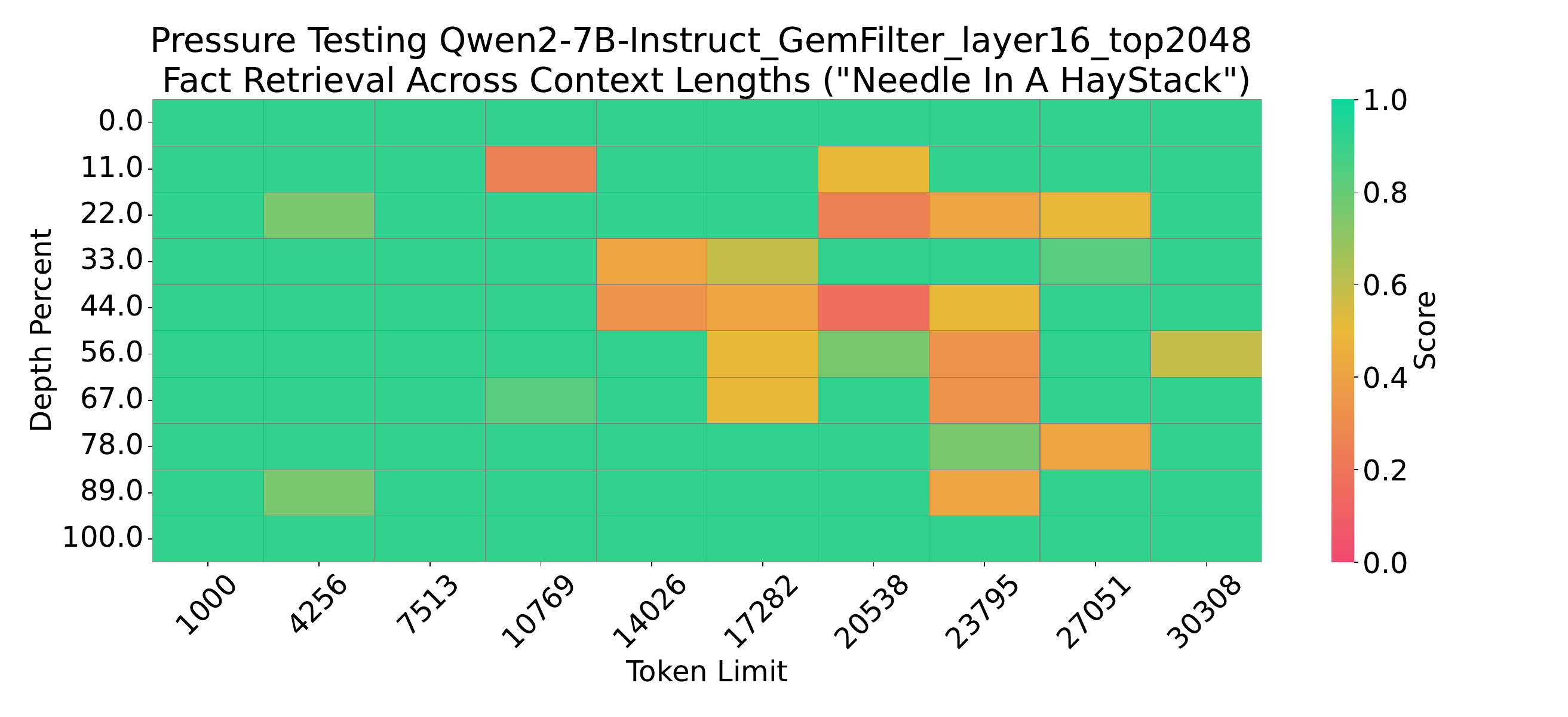}
            \label{fig:qwengem2048}
        }
        \subfigure[LLama 3.1 8B Instruct + SnapKV-2048: Score = 71.7]{
            \includegraphics[width=.46\textwidth]{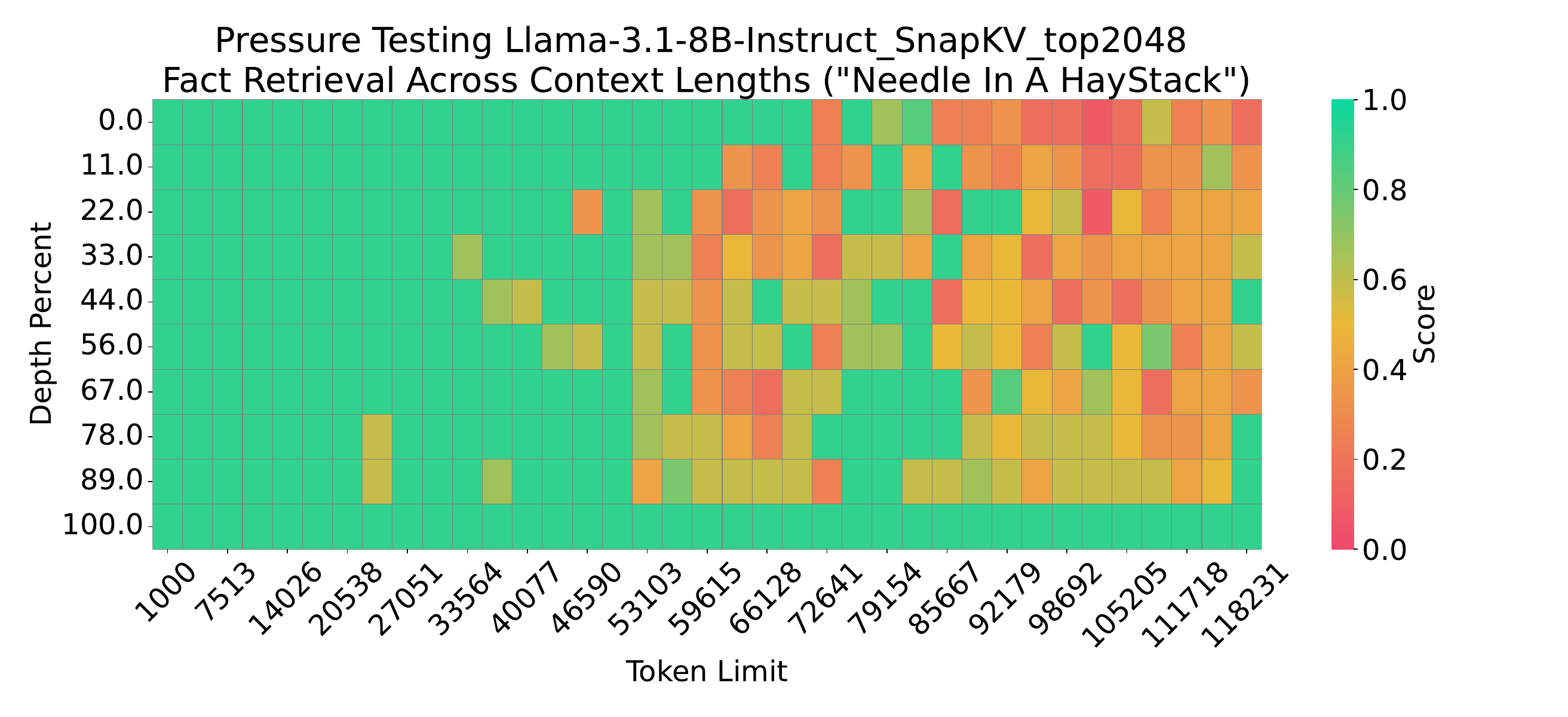}
            \label{fig:llamasnap2048}
        }
        \subfigure[Qwen2 7B Instruct + SnapKV-2048: Score = 90.8]{
            \includegraphics[width=.46\textwidth]{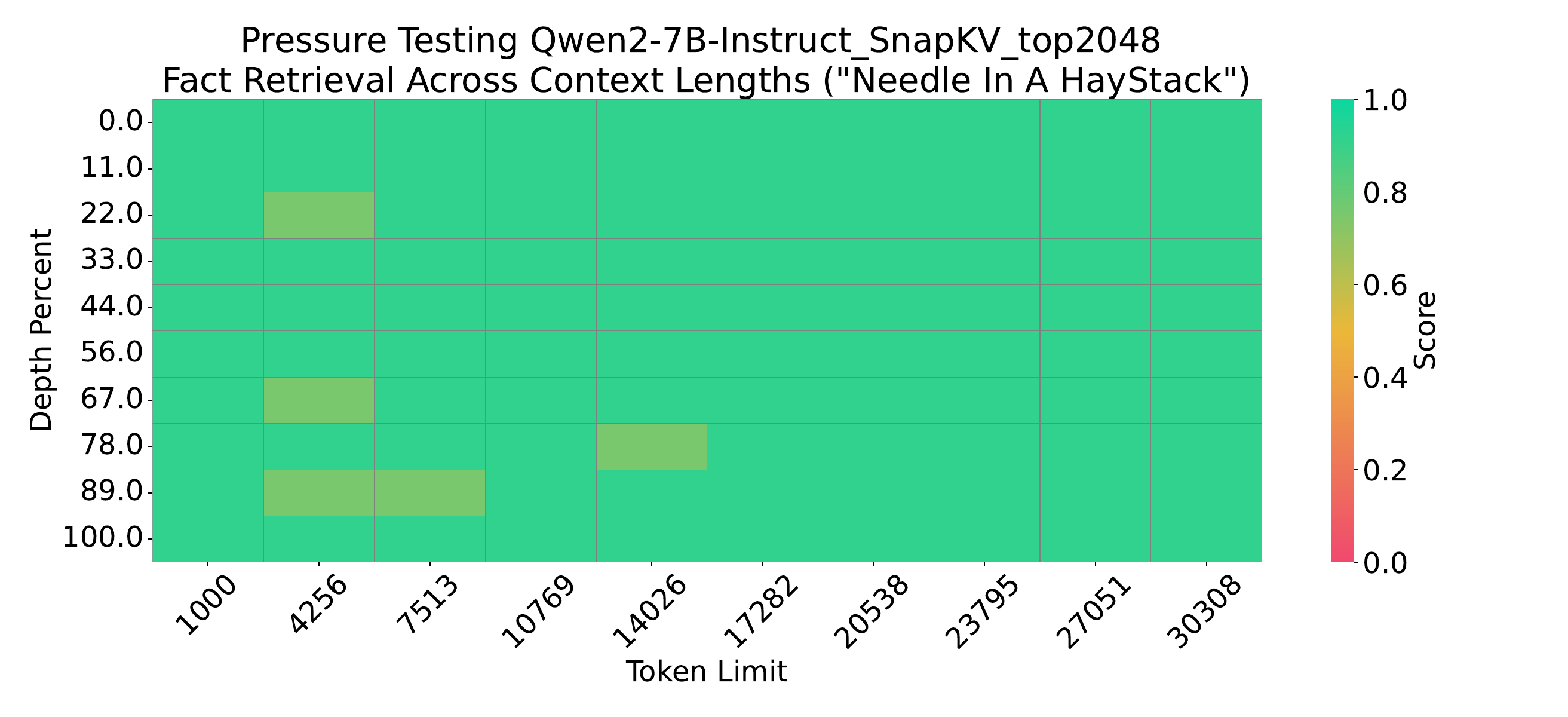}
            \label{fig:qwensnap2048}
        }
        \caption{Evaluation Results for Needle in a Haystack: LLaMA 3.1 8B Instruct's and Qwen2 7B Instruct's scores in all test cases with model settings PromptDistill-2048, GemFilter-2048, SnapKV-2048. (The x-axis represents the length of the input tokens, while the y-axis shows the position depth percentage of the ‘needle’ information)}
    \end{center}
    \label{fig:needle2}
\end{figure*}

\begin{figure*}[ht]
    \begin{center}
        \subfigure[LLama 3.1 8B Instruct + PromptDistill-4096: Score = 84.6]{
            \includegraphics[width=.46\textwidth]{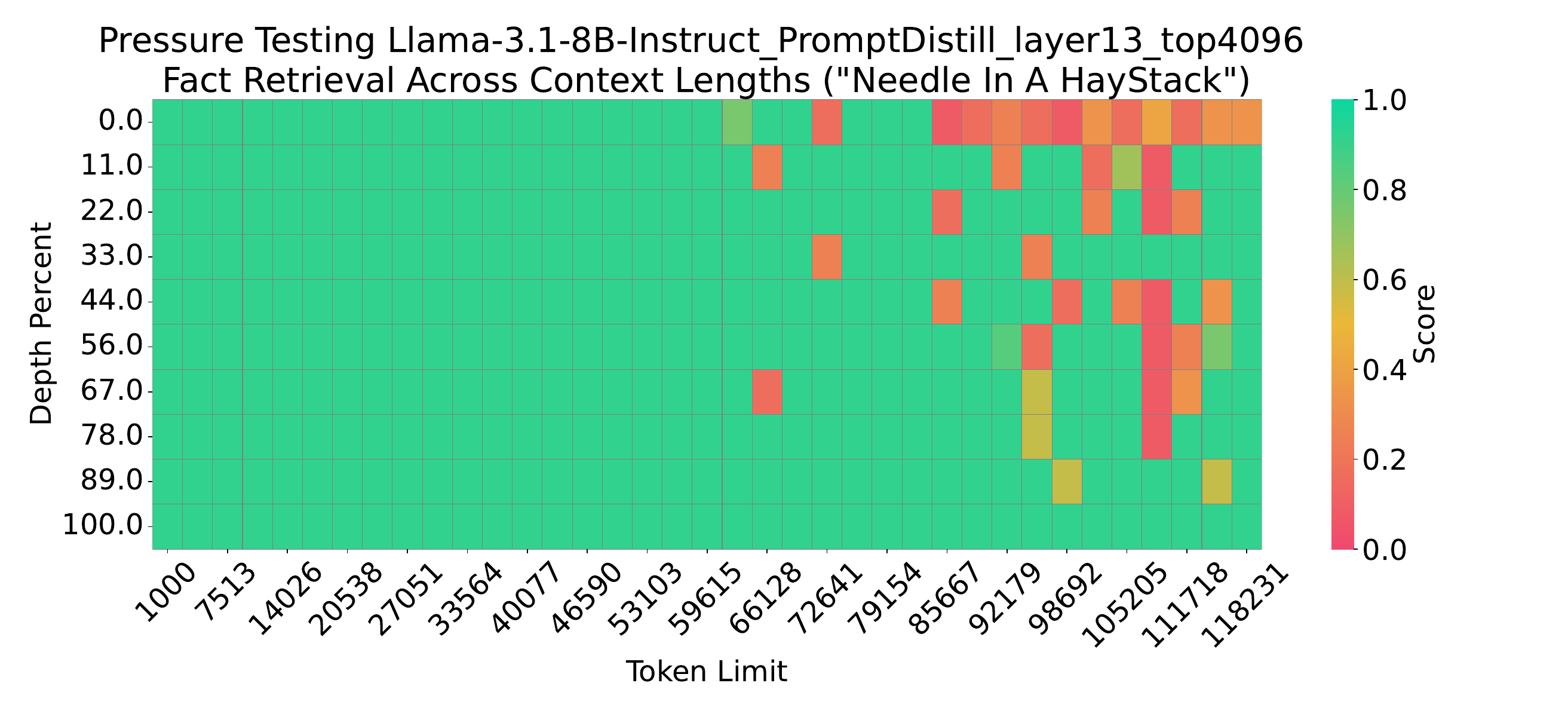}
            \label{fig:llamadist4096}
        }
        \subfigure[Qwen2 7B Instruct + PromptDistill-4096: Score = 90.2]{
            \includegraphics[width=.46\textwidth]{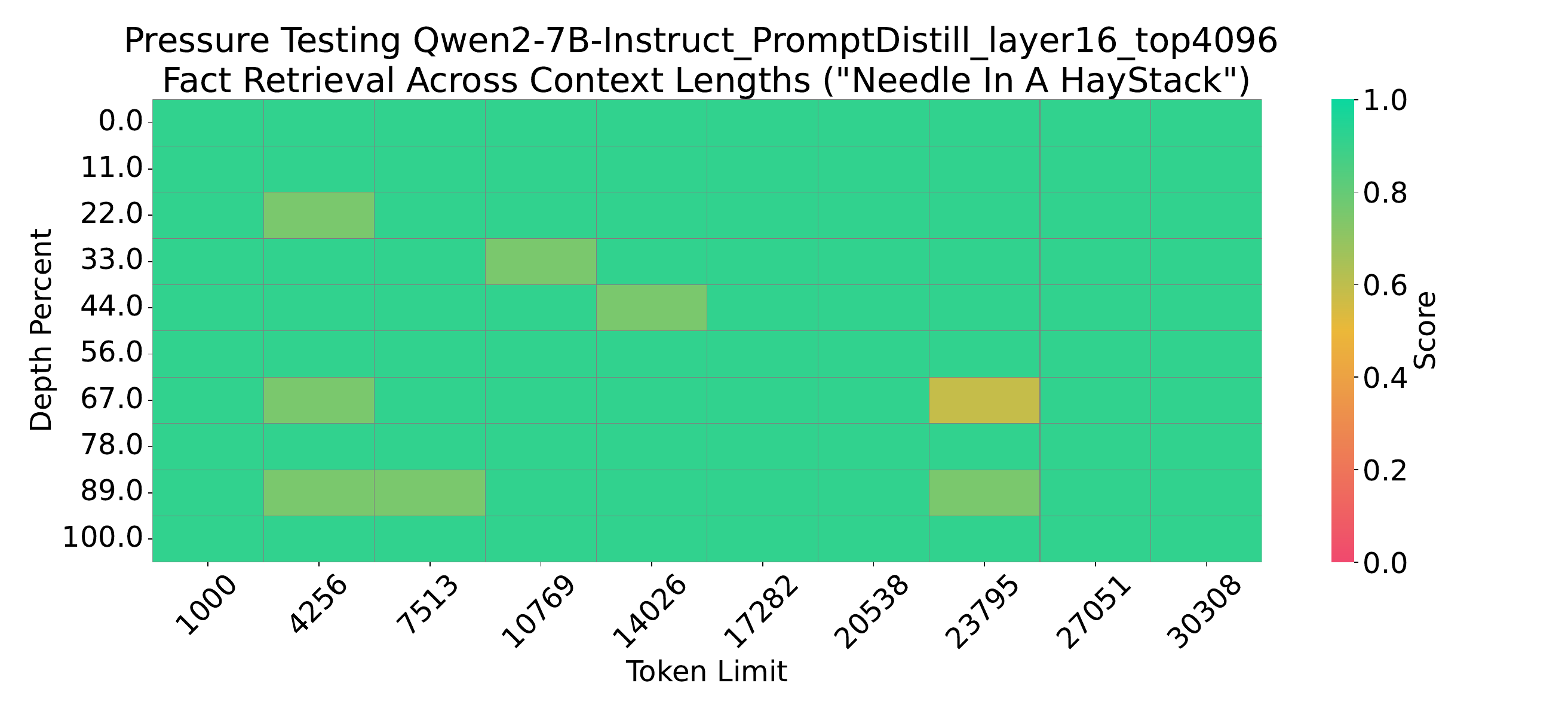}
            \label{fig:qwendist4096}
        }
        \subfigure[LLama 3.1 8B Instruct + GemFilter-4096: Score = 83.3]{
            \includegraphics[width=.46\textwidth]{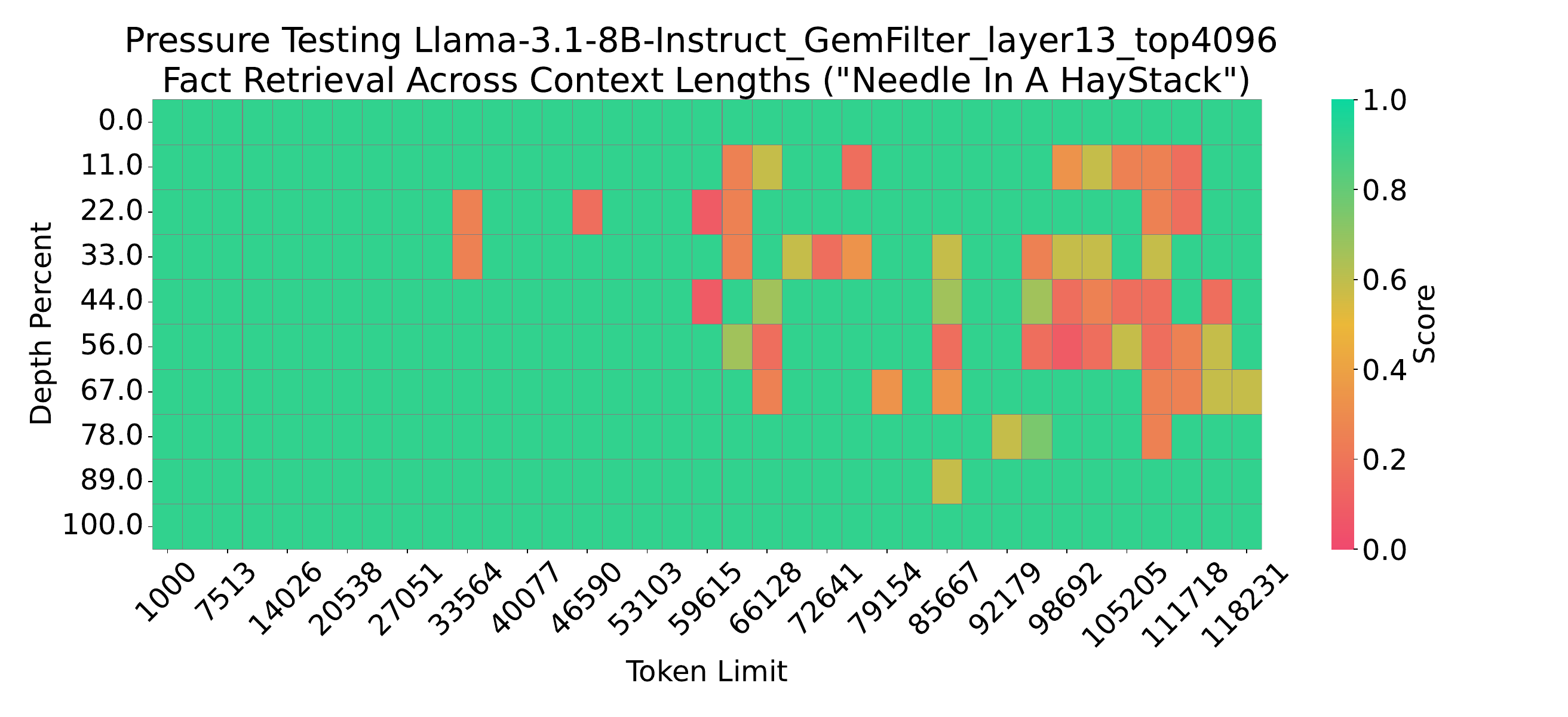}
            \label{fig:llamagem4096}
        }
        \subfigure[Qwen2 7B Instruct + GemFilter-4096: Score = 89.3]{
            \includegraphics[width=.46\textwidth]{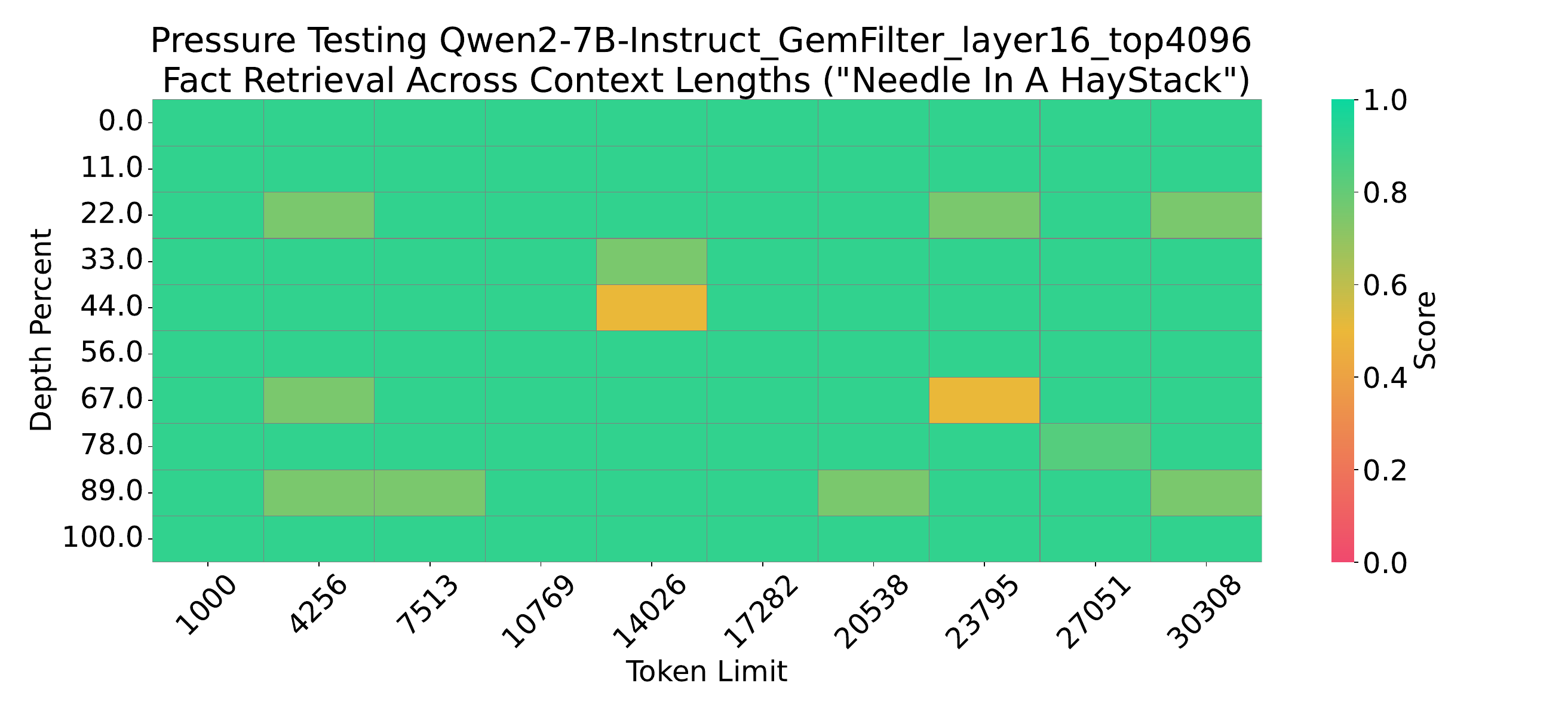}
            \label{fig:qwengem4096}
        }
        \subfigure[LLama 3.1 8B Instruct + SnapKV-4096: Score = 77.1]{
            \includegraphics[width=.46\textwidth]{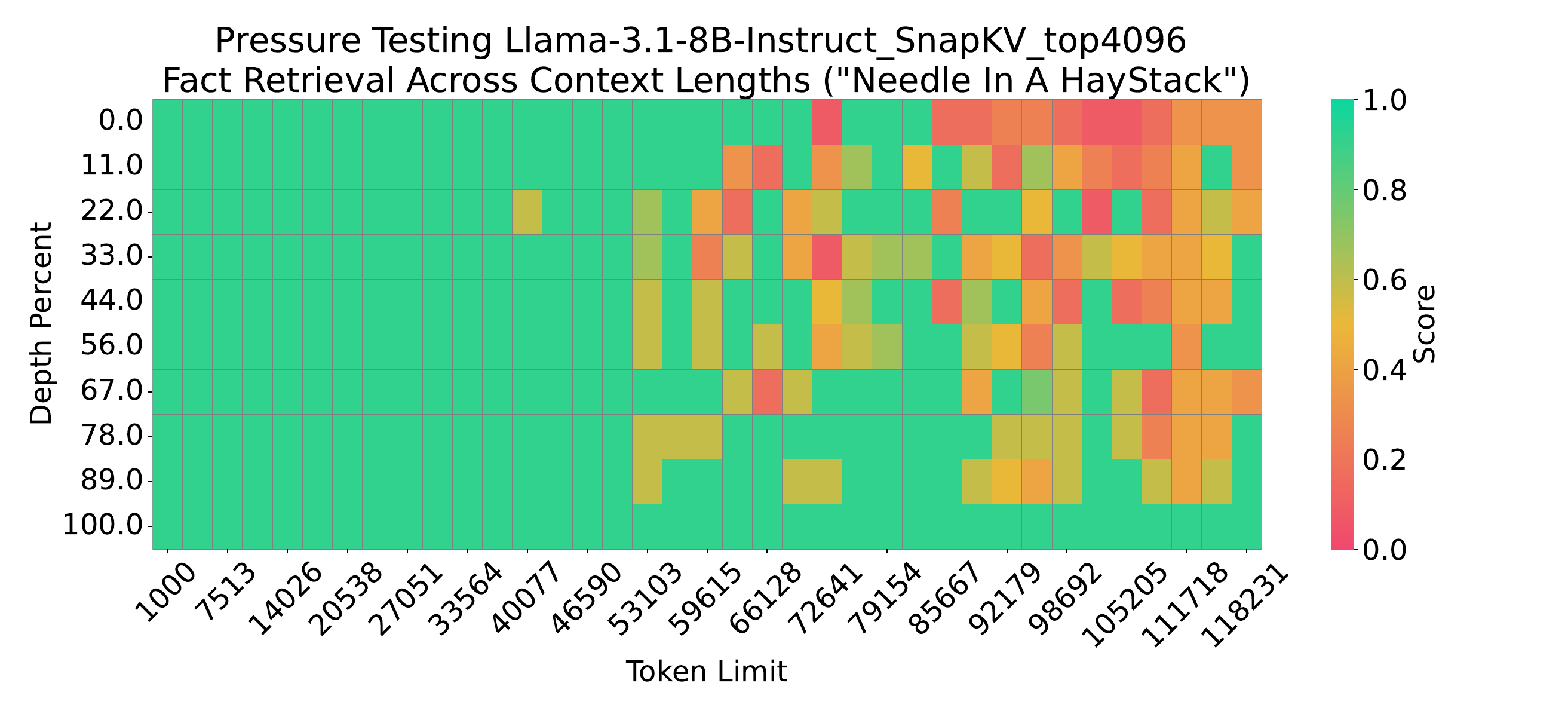}
            \label{fig:llamasnap4096}
        }
        \subfigure[Qwen2 7B Instruct + SnapKV-4096: Score = 91.0]{
            \includegraphics[width=.46\textwidth]{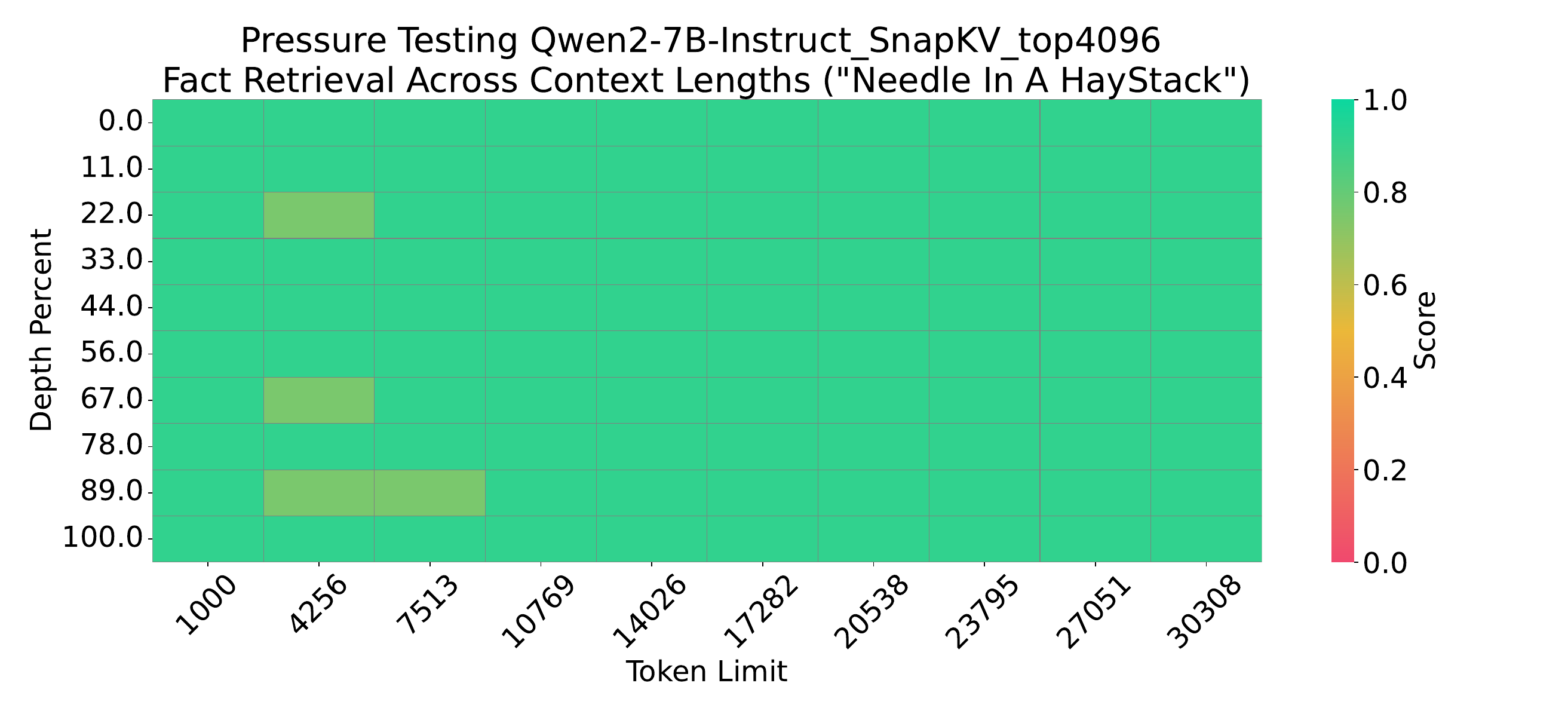}
            \label{fig:qwensnap4096}
        }
        \caption{Evaluation Results for Needle in a Haystack: LLaMA 3.1 8B Instruct's and Qwen2 7B Instruct's scores in all test cases with model settings PromptDistill-4096, GemFilter-4096, SnapKV-4096. (The x-axis represents the length of the input tokens, while the y-axis shows the position depth percentage of the ‘needle’ information)}
    \end{center}
    \label{fig:needle3}
\end{figure*}

\begin{figure*}[ht]
    \begin{center}
        \subfigure[Phi 3.5 Mini 3.8B Instruct + All KV: Score = 91.7]{
            \includegraphics[width=.46\textwidth]{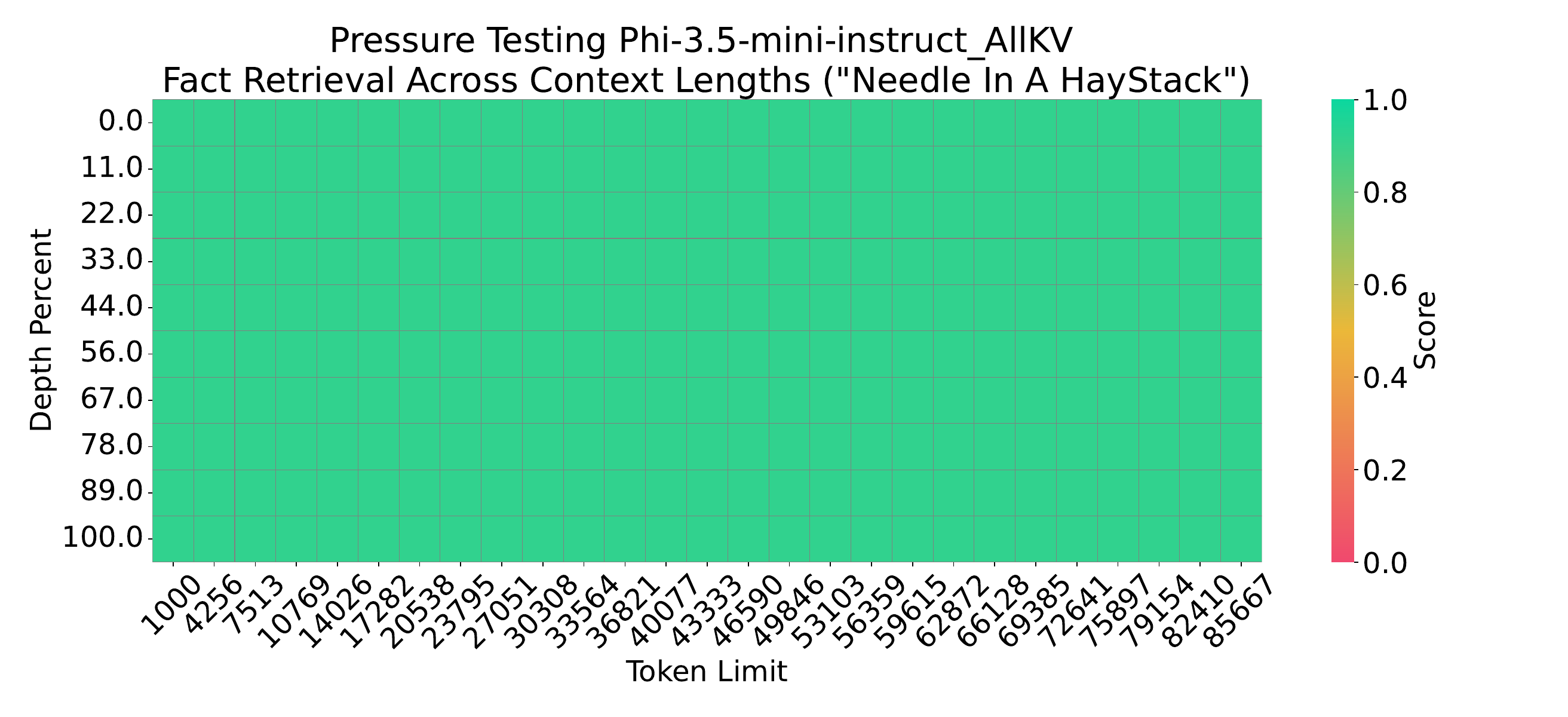}
            \label{fig:phidefault}
        }
        \subfigure[Phi 3.5 Mini 3.8B Instruct + PromptDistill-1024: Score = 91.7]{
            \includegraphics[width=.46\textwidth]{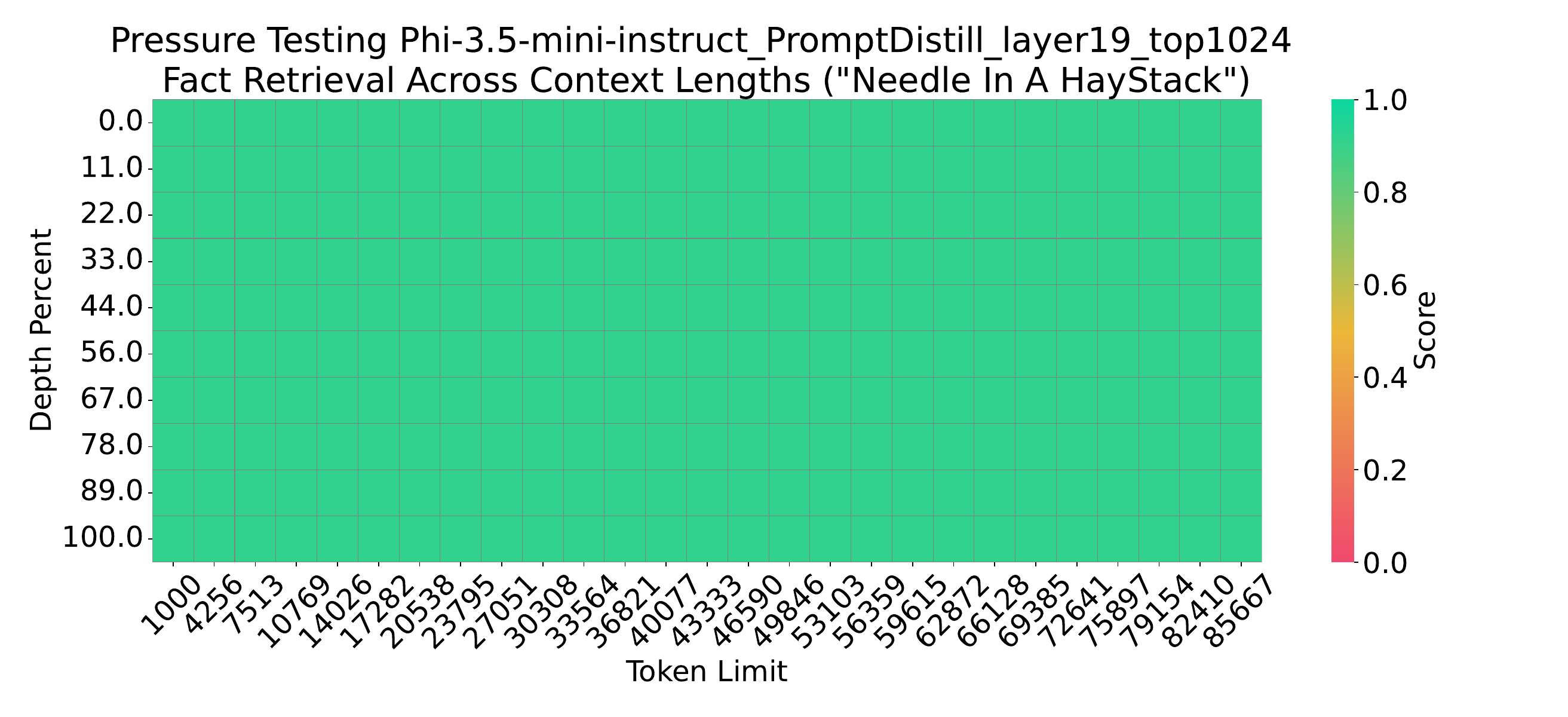}
            \label{fig:phidist1024}
        }
        \subfigure[Phi 3.5 Mini 3.8B Instruct + GemFilter-1024: Score = 91.7]{
            \includegraphics[width=.46\textwidth]{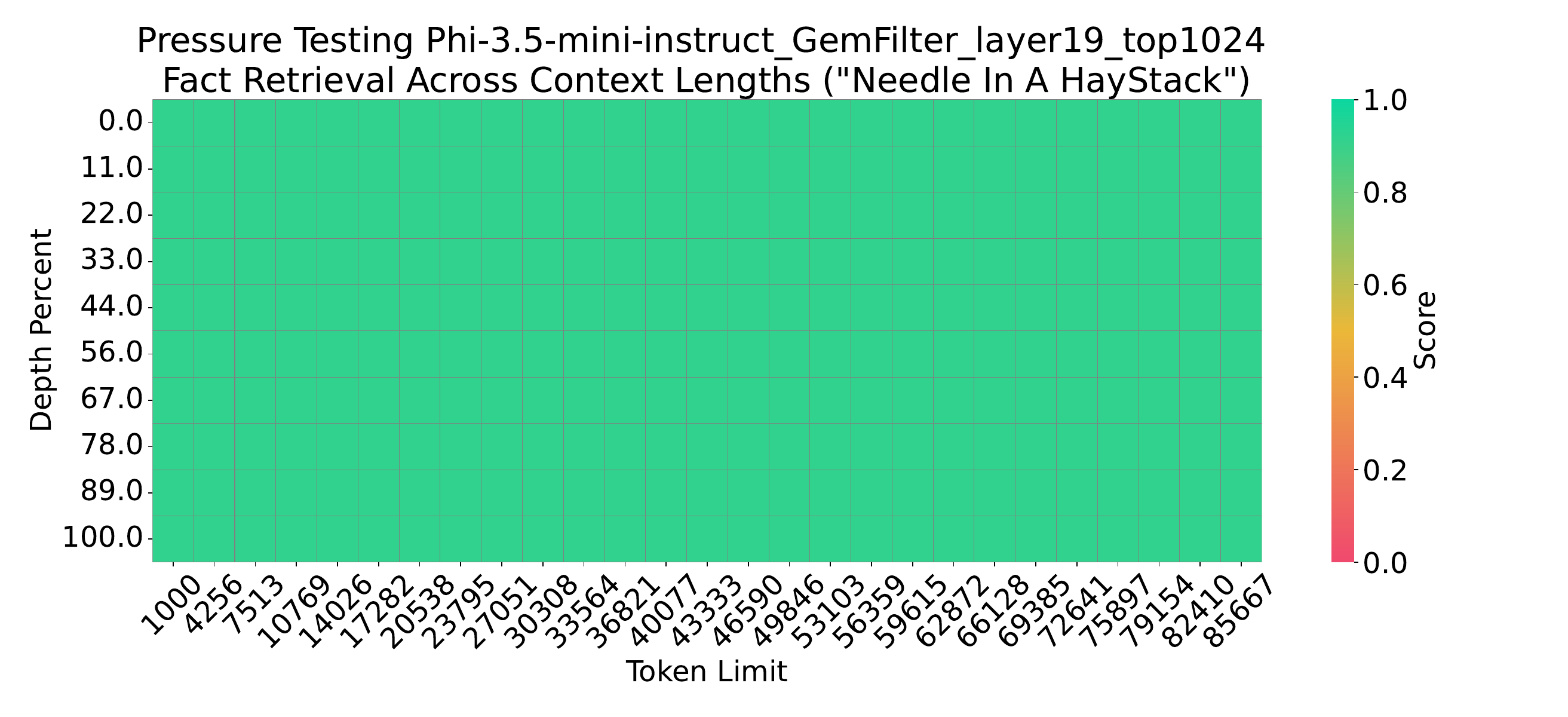}
            \label{fig:phigem1024}
        }
        \subfigure[Phi 3.5 Mini 3.8B Instruct + SnapKV-1024: Score = 91.7]{
            \includegraphics[width=.46\textwidth]{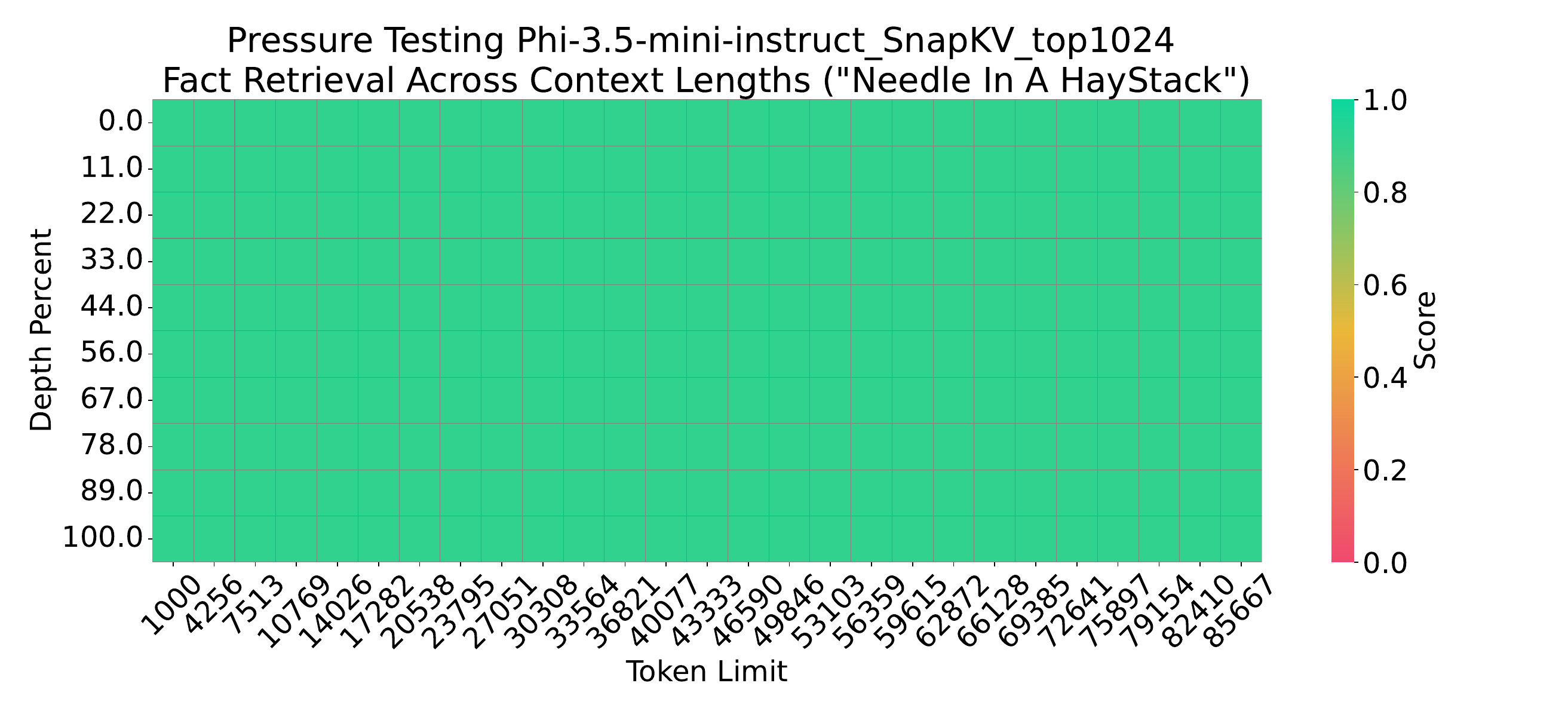}
            \label{fig:phisnap1024}
        }
        \subfigure[Phi 3.5 Mini 3.8B Instruct + PromptDistill-2048: Score = 91.7]{
            \includegraphics[width=.46\textwidth]{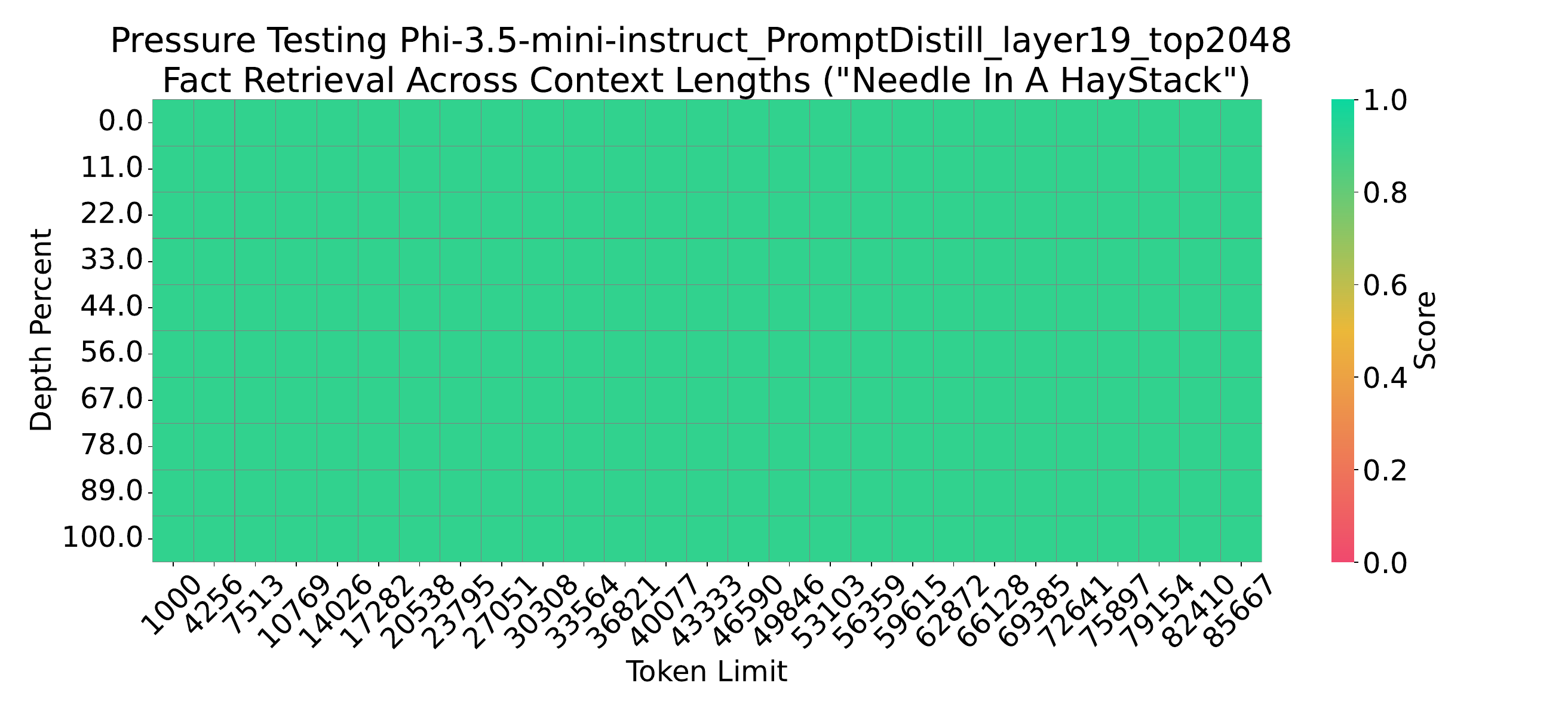}
            \label{fig:phidist2048}
        }
        \subfigure[Phi 3.5 Mini 3.8B Instruct + GemFilter-2048: Score = 91.7]{
            \includegraphics[width=.46\textwidth]{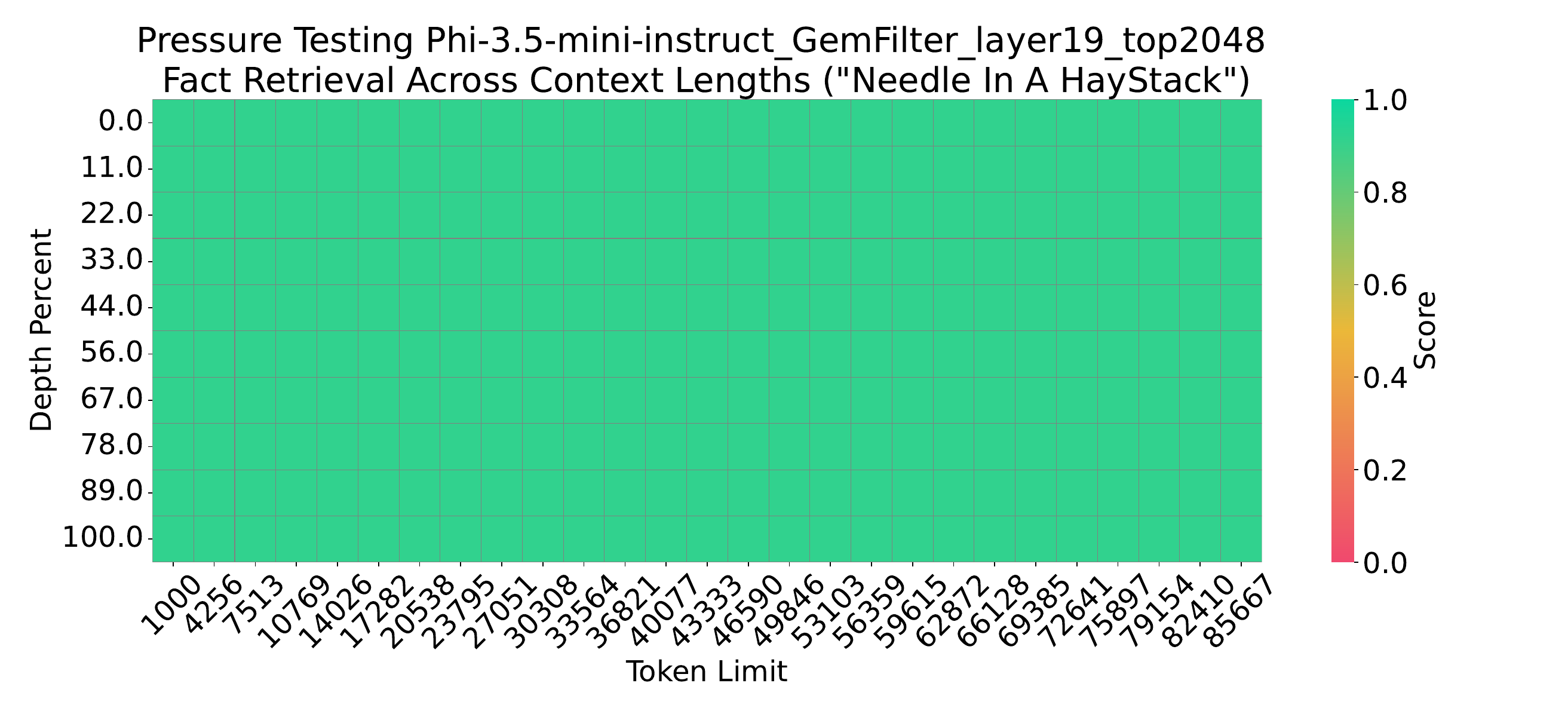}
            \label{fig:phigem2048}
        }
        \subfigure[Phi 3.5 Mini 3.8B Instruct + SnapKV-2048: Score = 91.7]{
            \includegraphics[width=.46\textwidth]{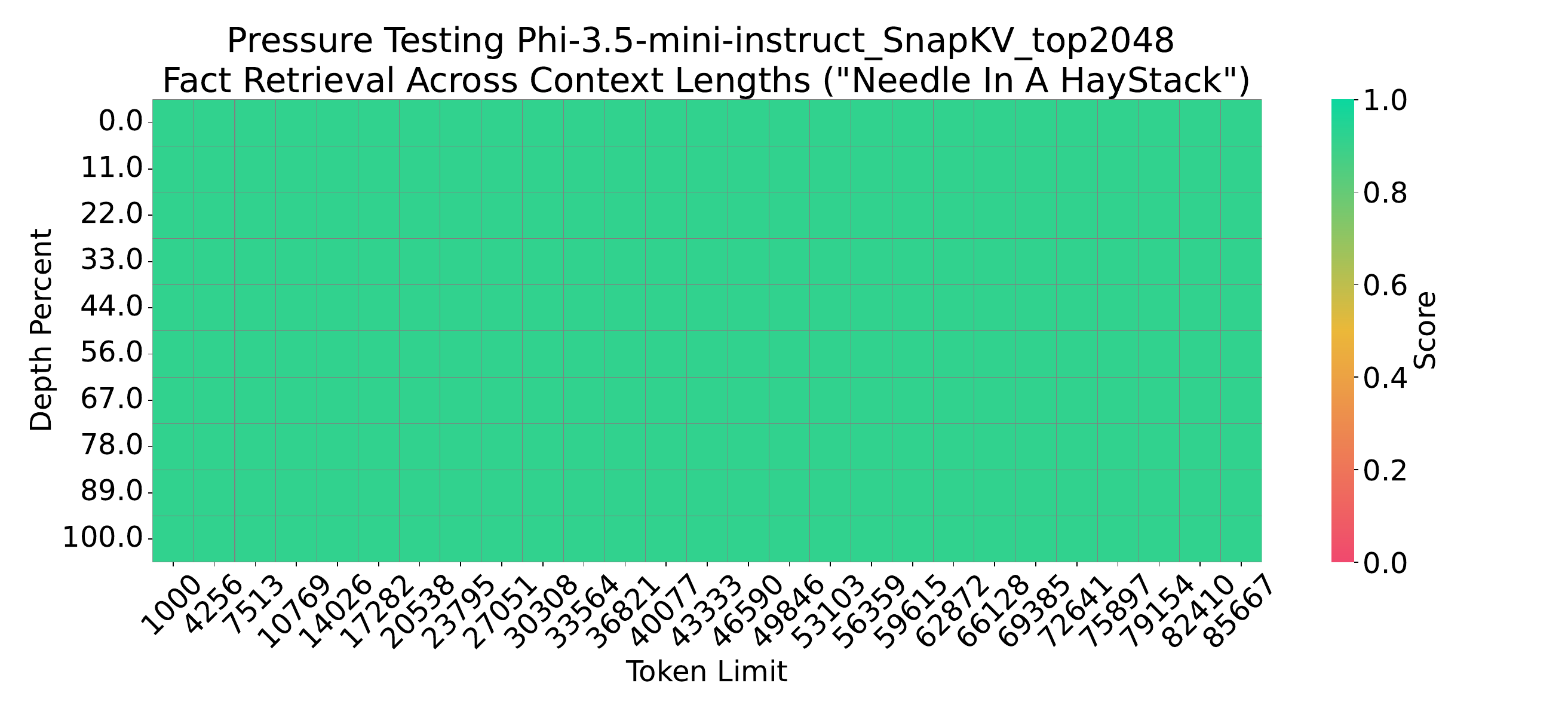}
            \label{fig:phisnap2048}
        }
        \subfigure[Phi 3.5 Mini 3.8B Instruct + PromptDistill-4096: Score = 91.7]{
            \includegraphics[width=.46\textwidth]{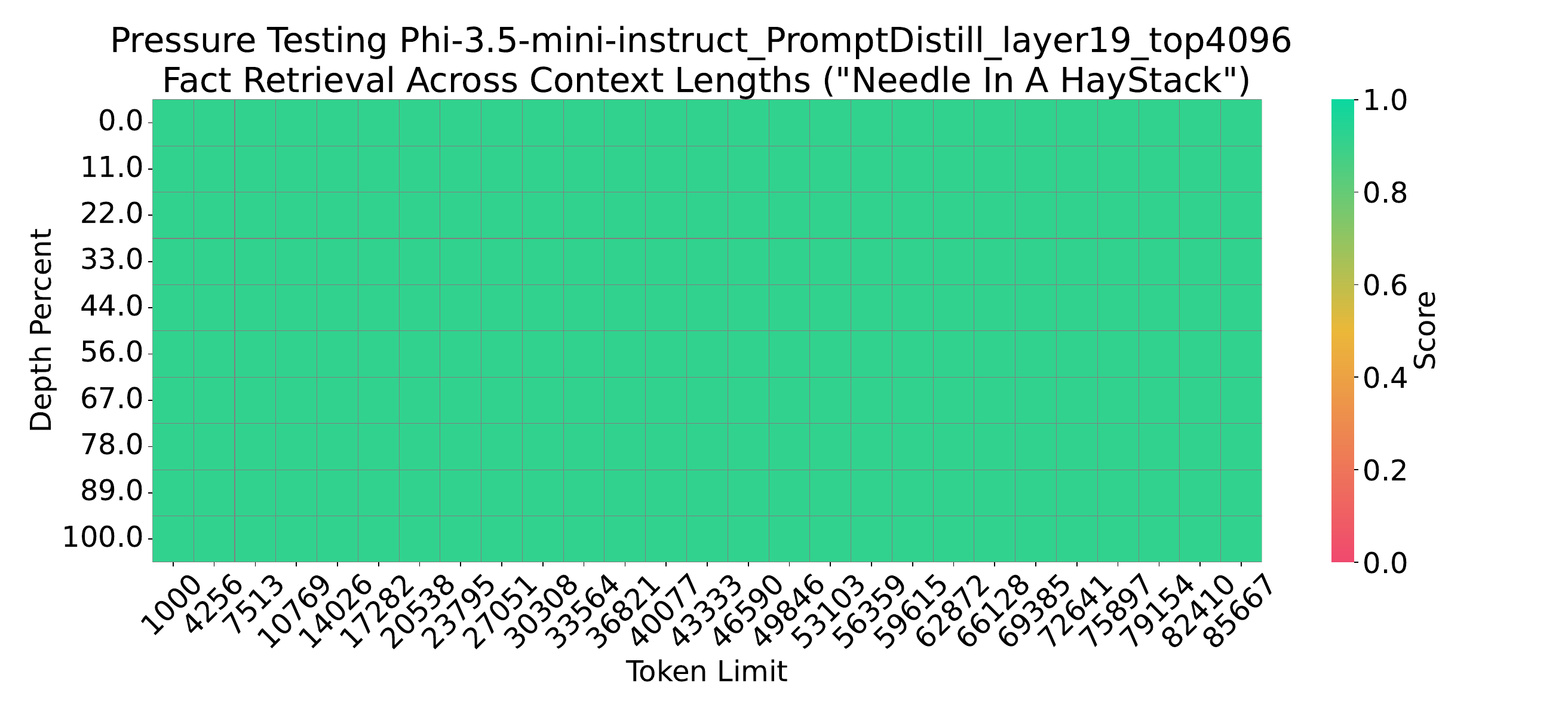}
            \label{fig:phidist4096}
        }
        \subfigure[Phi 3.5 Mini 3.8B Instruct + GemFilter-4096: Score = 5.1]{
            \includegraphics[width=.46\textwidth]{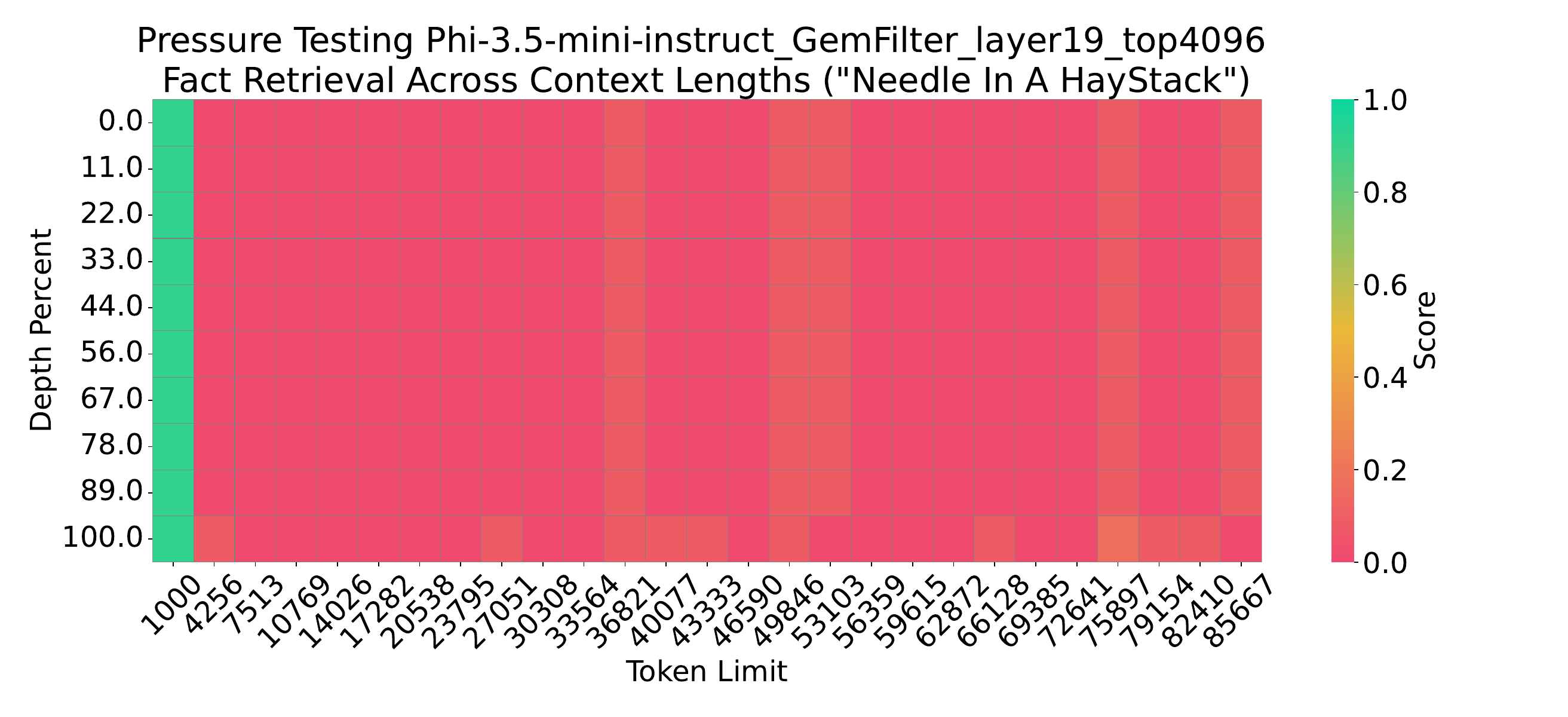}
            \label{fig:phigem4096}
        }
        \subfigure[Phi 3.5 Mini 3.8B Instruct + SnapKV-4096: Score = 91.7]{
            \includegraphics[width=.46\textwidth]{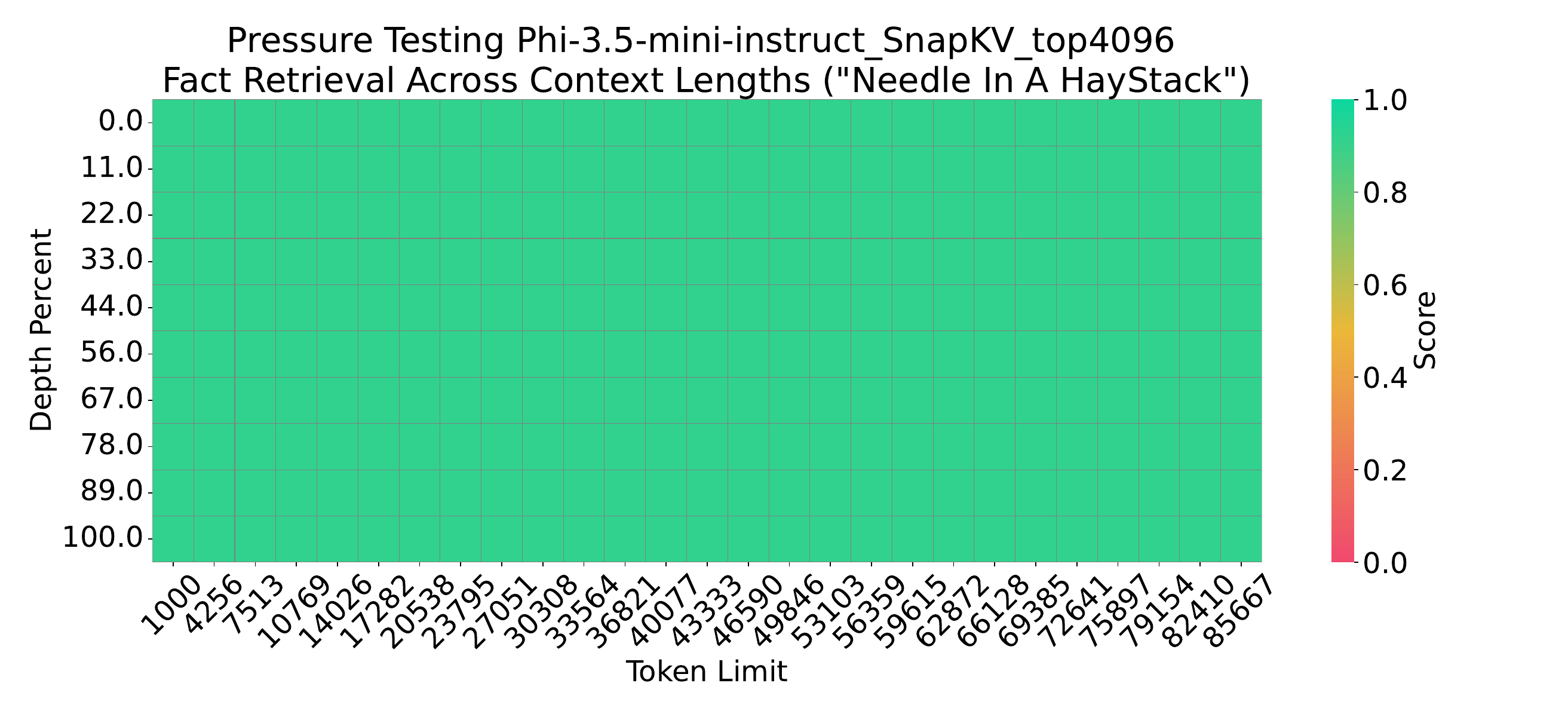}
            \label{fig:phisnap4096}
        }
        \caption{Evaluation Results for Needle in a Haystack: Phi 3.5 Mini 3.8B Instruct's scores in all test cases with all model settings. (The x-axis represents the length of the input tokens, while the y-axis shows the position depth percentage of the ‘needle’ information)}
    \end{center}
    \label{fig:needle4}
\end{figure*}

\end{document}